\documentclass[journal]{IEEEtran}
\usepackage{graphicx}
\usepackage{amsmath}
\usepackage[T1]{fontenc}
\usepackage[normalem]{ulem}%
\usepackage[latin9]{inputenc}
\usepackage{float}
\usepackage{bm}
\usepackage{mathrsfs}
\usepackage{subfigure}
\usepackage{amssymb}
\usepackage{environ}
\usepackage{rotating}
\usepackage{yhmath}
\usepackage{graphicx}
\usepackage{subfigure}
\usepackage{multirow}
\usepackage{picins}
\usepackage{multicol}
\usepackage{ulem}
\usepackage{color} 
\usepackage{colortbl}
\usepackage{cite}
\usepackage[ruled,vlined,linesnumbered]{algorithm2e}
\usepackage{booktabs}
\usepackage{makecell}
\graphicspath{{figures/}}

\definecolor{mydeepgray}{gray}{.4}
\definecolor{mygray}{gray}{.6}
\usepackage[pdftex]{hyperref}
\usepackage{lineno,hyperref}  
\modulolinenumbers[5]
\graphicspath{{figures/}}
\usepackage{graphicx}
\usepackage{environ}
\NewEnviron{myequation}{%
\begin{equation}
\scalebox{0.8}{$\BODY$}
\end{equation}
}

\NewEnviron{myequation1}{%
\begin{equation}
\scalebox{0.9}{$\BODY$}
\end{equation}
}

\begin{document}

\title{Pareto-wise Ranking Classifier for Multi-objective Evolutionary Neural Architecture Search}

\author{Lianbo~Ma,~\IEEEmembership{Senior Member,~IEEE,}
        Nan~Li,
        Guo Yu,~\IEEEmembership{Member,~IEEE,}
        Xiaoyu Geng,
        Shi Cheng,
        \\Xingwei Wang,
        Min Huang,
        Yaochu~Jin,~\IEEEmembership{Fellow,~IEEE}
\thanks{This work is supported by the NSFC Key Supported Project of the Major Research Plan Grant (No. 92267206), the National Natural Science Foundation of China (No.62032013), the National Key R\&D Program of China (No. 2022YFB4500800), the National Natural Science Foundation of China (No. 62136003, 62103150), the Joint Funds of the Natural Science Foundation of Liaoning Province (No. 2021-KF-11-01), the Fundamental Research Funds for the Central Universities (No.N2117005), and the China Postdoctoral Science Foundation (No. 2021M691012). (\textit{Corresponding authors: Guo Yu; Xingwei Wang; Min Huang})}

\thanks{Lianbo Ma, Nan Li, and Xiaoyu Geng are from College of Software, Northeastern University, Shenyang 110819, China. (email: malb@swc.neu.edu.cn, lnnner@163.com, and 2071277@stu.neu.edu.cn)} 
\thanks{Guo Yu is with the Institute of Intelligent Manufacturing, Nanjing Tech University, Nanjing, 211816, China.  (email: gysearch@163.com).}
\thanks{Shi Cheng is with the School of Computer Science, Shaanxi Normal University, Xi'an, 710062 China (e-mail: cheng@snnu.edu.cn).}
\thanks{Xingwei Wang is with the College of Computer Science, Northeastern	University, Shenyang 110819, China (e-mail: wangxw@mail.neu.edu.cn). }
\thanks{Min Huang is with the College of Information Science and Engineering, State Key Laboratory of Synthetical Automation for Process Industries, Northeastern University, Shenyang 110819, China (e-mail: mhuang@mail.neu.edu.cn).}
\thanks{Yaochu Jin is with the Faculty of Technology, Bielefeld University, 33619 Bielefeld, Germany. He is also with the Department of Computer Science, University of Surrey, Guildford, GU2 7XH, UK. (email: yaochu.jin@unibielefeld.de.)}
}

\maketitle

\begin{abstract}
In multi-objective evolutionary neural architecture search (NAS), existing predictor-based methods commonly suffer from the rank disorder issue that a candidate high-performance architecture may have a poor ranking compared with the worse architecture in terms of the trained predictor.To alleviate the above issue, we aim to train a Pareto-wise end-to-end ranking classifier to simplify the architecture search process by transforming the complex multi-objective NAS task into a simple classification task. To this end, a classifier-based Pareto evolution approach is proposed, where an online classifier is trained to directly predict the dominance relationship between the candidate and reference architectures. Besides, an adaptive clustering method is designed to select reference architectures for the classifier, and an $\alpha$-domination assisted approach is developed to address the imbalance issue of positive and negative samples. The proposed approach is compared with a number of state-of-the-art NAS methods on widely-used test datasets, and  computation results show that the proposed approach is able to alleviate the rank disorder issue and outperforms other methods. Especially, the proposed method is able to find a set of promising network architectures with different model sizes ranging from 2M to 5M under diverse objectives and constraints.
\end{abstract}

\begin{IEEEkeywords}
Neural architecture search, Pareto evolution, dominationship classification, multi-objective search. 
\end{IEEEkeywords}
 
\IEEEpeerreviewmaketitle

\section{Introduction}
\IEEEPARstart{N}{eural} architecture search (NAS) is known as an effective way to  automate the design of task-specific neural network  architectures, instead of designing hand-crafted DNNs based on extensive human expertise \cite{1r,2r,3r,4r}. It has been demonstrated that NAS can not only achieve competitive DNN architectures as human experts do, but also discover new state-of-the-art architectures \cite{3r,4r}. 

Starting from Google Brain's NAS \cite{13r}, the development of NAS has experienced two stages \cite{23r}. The early stage has seen the development of NAS based on training-from-scratch \cite{24r, 2rr, 25r, 26r}, such as NASNet \cite{14r} and NAS-RL \cite{13r}. They are nevertheless time-consuming and computationally expensive \cite{27r}. For example, NASNet \cite{14r} has to consume 32400-43200 GPU hours for the training on a common dataset like CIFAR-10. In the current stage, one-shot NAS methods based on weight sharing \cite{4r, 5r, 6r, 16r, 28r, 29r} are prevalent. That is, the parameters of candidate architectures are directly inherited from a supernet. In this way, we only need to train the supernet so that the performance evaluation can be significantly speeded up \cite{7r}, \cite{30r}. For example, Sinha et al. \cite{a6} improved the weight sharing approach with a new decoding technique of the architecture parameters, which can divert more gradient information for the given architectures, so that the predictive capability of the given architectures is enhanced by this way. Overall, the existing NAS approaches can be roughly divided into EA-based \cite{12r}, RL-based \cite{19r}, gradient-based \cite{16r}, Bayesian-based \cite{31r, 32r}, and local/random search-based \cite{29r} classes.

From the perspective of multi-objective optimization, NAS is a complex multi-objective optimization problem, which involves conflicting multiple objectives (e.g., accuracy, FLOPs, energy, and latency) \cite{5r,7r,13r}, and computationally expensive architecture evaluation \cite{6r,7r,8r}. To solve the multi-objective NAS, a variety of evolutionary algorithms NAS (ENAS) methods have been developed \cite{10r,11r,sun2020automatically, a2}. For example, Lu et al. \cite{a2} presented NSGANetV1, one of the most representative multi-objective methods to optimize CNN architectures, where the classification performance and FLOPs are taken as two conflicting objectives. In this study, they used the Bayesian models and downsampling architectures to improve the computational efficiency. In addition, Dudziak et al. \cite{a3} aimed to find the hardware-aware architectures by optimizing two objectives (i.e., latency and accuracy), where the latency is replaced by a graph convolutional network (GCN) based latency predictor. In contrast to the conventional reinforcement learning (RL) \cite{5r, 13r}, and gradient algorithms \cite{16r, 49r} driven methods, EA-based NAS (ENAS) methods \cite{9r,10r,11r,12r} are free from the gradient information of the objectives and they are insensitive to the discontinuity and non-differentiability of the objectives \cite{sun2018igd,13r,14r,15r}.

The popular way of existing ENAS methods is to utilize performance predictors to predict the performance value of each candidate, which can largely save the expensive costs of architecture evaluation. For example, PRE-NAS \cite{a7} selected a small amount of representative architectures as the training samples for online accuracy predictor, where the representative architectures were evaluated by high-fidelity weight inheritance during the evolutionary search. The effectiveness of such strategy has been fully demonstrated in the extended work of PRE-NAS \cite{a8}. Typically, they follow the same procedures: (1) train a high-accurate surrogate (e.g., regression model) based on a set of well-trained architectures; (2) predict the numerical value of the accuracy of a candidate architecture; and (3) determine the optimal solutions according to the numerical values of multiple objectives. However, these procedures easily bear the \emph{rank disorder issue} \cite{3r} that a candidate architecture with high-performance may have a poor ranking compared with the worse architecture according to the trained predictor. The reasons are as follows.

(1) The surrogate model fitting the loss function (e.g. MSE) may easily lead to the rank disorder issue \cite{7r,8r}. Taking MSE as an example, we compare two architectures $A$ and $B$, whose ground-truth accuracy is 0.8 and 0.9, respectively. Due to the squared residuals \cite{3r}, we may get a reasonable result from the model that the predicted accuracy of them can be 0.85 for $A$ and 0.84 for $B$. Thereby, $A$ with higher prediction accuracy will be selected although $B$ is better in practice. Consequently, the search can be misled by the solutions with low accuracy but they are mispredicted with high accuracy.

(2) The limited computing resources or expensive evaluations make the rank disorder issue even worse during the architecture search \cite{16r,17r}. The reason is that a highly accurate regression model is difficult to get under the limit of training samples, so that inaccurate performance evaluations will be performed and further deteriorate the {rank disorder issue}. Once the {rank disorder issue} occurs, the Pareo dominance relationship tends to be ineffective, especially when the number of objectives increases \cite{7r}\cite{21r}.

To alleviate the \emph{rank disorder issue} and build robust surrogate model under the limited amount of computing resources, we are inspired by the ordinal optimization theory \cite{22r} and come up with an idea, only qualitatively determining the architectures whether they are good or not, instead of quantificationally comparing them with the predicted objective values or performance indicators. In fact, our idea is in accord with the end-to-end nature of ENAS. That is, the essential goal of the search in ENAS is to select a number of promising architectures with good performance into the next-round evolution during the environmental selection.

Accordingly, we propose a \textbf{C}omparator-based Pareto \textbf{E}volution approach for \textbf{NAS} (CENAS), where an online Pareto-wise end-to-end ranking classifier is learnt to recognize optimal architectures in the multi-objective search. Such classifier can predict the dominant relationship between the candidate architecture and the reference architectures selected by an adaptive clustering method. In other words, the candidate architectures are classified into good and poor classes, where the good architectures either can dominate or are non-dominant with the reference ones, and the poor ones are dominated by the references. To avoid the class imbalance problem, we design an $\alpha$-dominance evaluation method to enhance the classification accuracy by regulating the rate between the candidate architectures in good class and poor class.

The main contributions of this paper include:

\begin{enumerate}
		
  \item The Pareto-wise end-to-end ranking classifier is designed to transform a complex multi-objective NAS task into a simple Pareto-dominance classification task. Such an ordinal optimization-based classifier can effectively alleviate the problem of rank disorder in environmental selection. This is the first attempt to utilize the Pareto dominance classification concept to simplify the search process of NAS.
  
  \item The classifier-based Pareto evolution framework is devised, with the aim of enhancing the efficiency and effectiveness of NAS. In addition, two adaptive schemes are designed to ensure the reliability of the classifier during the evolutionary process. The one is to select reference architectures for the designed classification, while the other is to eliminate the class imbalance issue by $\alpha$-dominance evaluation.
  
  \item The experimental results demonstrate the practicality and scalability of the proposed framework on multiple widely-used datasets, and achieves cutting-edge results in various critical performance metrics (especially search efficiency). Besides, we further show the performance of the proposed method on Photovoltaic cell and achieves the best-known results.
\end{enumerate}

The rest of this paper is organized as follows. Section \ref{S-II} presents the motivation. Section \ref{S-III} details the proposed CENAS. In Section \ref{S-IV}, we present the experimental settings and comparative results between CENAS and other state-of-the-art approaches. Finally, the conclusions and future directions are drawn in Section \ref{S-V}.


\section{Motivation}\label{S-II} 

\begin{figure*}[htbp]
\centering
\includegraphics[width=\linewidth]{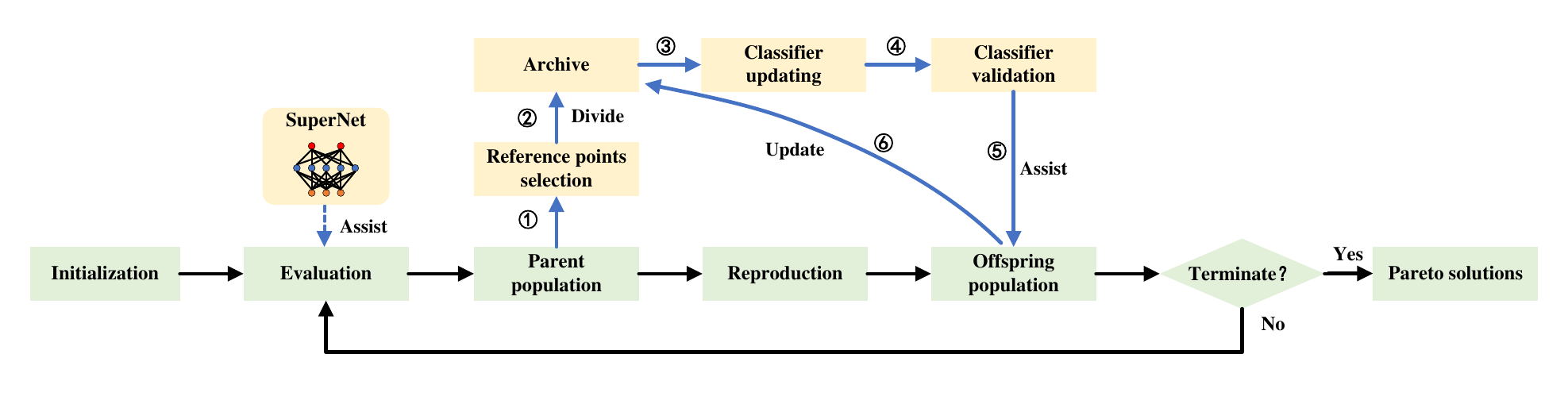}\\

\caption{An illustration of the framework of CENAS, where the numbers indicate the sequence of the operations.}\label{Fig1}

\end{figure*}

Existing ENAS methods based on performance predictors easily suffer from the rank disorder issue that a candidate high-performance architecture may have a poor ranking compared with the worse architecture. The reasons are as follows. First, the commonly used regression model fitting the loss function is to minimize the MSE, which may easily result in the rank disorder issue \cite{3r}. Due to the squared residuals, the predicted order of the candidate architectures is inconsistent with the real scenario, which ultimately misleads the search of ENAS. Unfortunately, it is not just the case for MSE but other loss functions such as Cross-entropy. Second, the limited computing resources or expensive evaluations make the rank disorder issue even worse. During the ENAS, we cannot provide with large numbers of architectures for model training. It is inevitable that inaccurate regression models will be achieved, which may present misleading predictions or evaluations. This is more serious when the number of objectives increases \cite{7r}.

Therefore, how to alleviate the rank disorder issue under the limited amount of computing resources is the key of ENAS methods to effectively and efficiently to choose the architectures with competitive performance during the environmental selection. We have found that the performance of EA-based NAS methods can be enhanced if the candidate architectures can be properly classified into ``good'' and ``poor'' classes. To this end, we will borrow the idea of ordinal optimization theory \cite{22r} to build an end-to-end Pareto classifier to transform the complex multi-objective NAS into a simple Pareto-dominance classification task. There are two basic principles of the ordinal optimization theory.

{\textbf{Claim 1:}} ``Order is easier than value''\cite{22r}. That is, it is easier to determine whether $A>B$ than to determine $A-B=?$ (or $A=?$ and $B=?$). Different from other approaches that predict the accuracy of an architecture (i.e., the value), we focus on whether the architecture is good enough to survive into next evolution (i.e., the order). This way can simplify the search process and enhance the search efficiency.

{\textbf{Claim 2:}} ``Nothing but the best is very costly''\cite{22r}. Finding the ``best'' architecture in each generation requires a ranking based on model accuracy, which is expensive through accuracy predictor since a large number of training samples are required to train the complex model to fit architecture and accuracy mapping function. Instead of finding the ``best'' architectures,  training a simple classifier to identify whether an architecture is  ``good'' or ``bad'' may largely reduce the computational cost.

Recent work \cite{a4} learned a neural architecture comparator to measure the superiority and inferiority of different architectures in terms of their accuracy. However, this study focuses on the comparison of their accuracy between different architectures so that this research cannot be directly extended to the multi-objective NAS. In handling multi-objective optimization, Pan et al. \cite{21r} fit the expensive objectives with the computationally cheap surrogate models, and replaced the Pareto dominance relationship with a classification surrogate model to predict a candidate solution whether it is dominated or non-dominated. Huang et al. \cite{a1} transformed a complex multi-objective task into a binary classification task and achieved promising experimental results, where the contrastive learning model was used to identify the dominant relationship between the two solutions. Inspired by the study \cite{21r}, \cite{a1}, we can extend the idea of classification-assisted multi-objective optimization to handle NAS.  

\section{Proposed method}\label{S-III}

\subsection{Main framework}
Following the conventional framework of multi-objective optimization \cite{67r}, the main framework of CENAS is shown in Fig. \ref{Fig1}. In order to alleviate the rank disorder issue and build robust surrogate model under the limited amount of computing resources, we train an online Pareto-wise end-to-end ranking classifier on the basis of a constructed archive, where the candidate architectures are classified into ``good'' or ``poor'' classes. As a result, it enables the environmental selection to select competitive candidate architectures without true evaluation either on real objectives or surrogates of objectives. In other words, a complex multi-objective NAS task is transformed into a simple Pareto-dominance classification task. In addition, the complexity of the algorithm is provided from time and space perspectives in Section X of the Supplementary material. The pseudo code of the proposed CENAS is presented in Algorithm \ref{Algori1} and the main steps are as follows:

\subsubsection{Initialization (Lines 1--4 of Algorithm \ref{Algori1})} Following the encoding of the search space (in Section III-B),  a population with $N$ architectures (subnets) is initialized by uniform sampling \cite{15r}, in which each possible operation is trained by the supernet with the same probability (1/8)\footnote{There are eight operations in this study, which are introduced in Table S1 of Section V of the Supplementary material. In addition, the number of operations and the types of operations are discussed in Section VIII of the Supplementary material.}\cite{12r}. Then, the subnets are evaluated on the validation set with respect to their objectives, where their corresponding weights are inherited from the supernet (in Line 2)\footnote{In traditional one-shot methods, it is known that one-shot model itself is not very reliable in terms of predicting true performance. By contrast, the performance evaluations of candidate architectures in our approach are divided into two stages: 1) inheriting weights from supernet; 2) fine-tuning the architecture to obtain accuracy. In this way, we can obtain their true performance under a small computational overhead.}. In the initialization stage, the labels of the population ($P$) are marked on the basis of $F$ (in Line 3), where 50\% of the population with higher accuracy is marked as ``good'' and the rest as ``poor''. Some other parameters are initialized in Line 4. 

\subsubsection{Selection of reference points (Line 6 of Algorithm \ref{Algori1})} This step is to select a number of reference architectures $S_R$ from population $P$ to construct the classification boundary. To this end, we develop an angle-based cluster strategy to select a set of reference points with the aim of adapting to the shape of the Pareto front. In addition, we utilize a relaxed dominance strategy to adaptively regulate the rate of samples in class 1 (``good'') and class 2 (``poor''), which will eliminate the class imbalance problem. The details of this step are shown in Algorithm \ref{Algori2}.

\subsubsection{Classifier updating and validation (Lines 7-10 of Algorithm \ref{Algori1})} On the basis of the classification boundary, the architectures in the archive are categorized into class 1 and class 2 and saved in $S_c$  in Line 7 (detailed in Algorithm \ref{Algori3}). Then, $S_c$ is divided into the training and validation data sets in Line 8 (detailed in Algorithm \ref{Algori4}), which are used for the training and validation of the proposed classifier, respectively (in Lines 9 and 10). In Line 10, $AUC$ defined in Eq. \ref{Eq6} indicates the reliability of a classifier, where the classifier is an ensemble of support vector machines (SVMs), termed as ESVM (see Algorithm \ref{Algori5}).

\subsubsection{Classifier-assisted environmental selection (Lines 11 and 12 of Algorithm \ref{Algori1})} In Line 11, reproduction operators are firstly performed on the parent solution set ($P\cup S_R$) to create an offspring solution set ($Q$). Then, elite architectures are selected from the solution set by the proposed classifier and saved into $P$ as the parent solution set for the next generation in Line 12 (detailed in Algorithm \ref{Algori6}), where the offspring solutions ($Q$) are evaluated with the supernet.

The key components of CENAS are detailed in the following subsections.

\begin{algorithm} 
	\DontPrintSemicolon
	\SetAlgoLined
	\SetKwInOut{Input}{Input}\SetKwInOut{Output}{Output}
	\Input{$N$: population size, $T$: number of iterations, $S_w$: supernet, $Th\_AUC$: predefined $AUC$ threshold, $S_t$: number of sampling.}
	
	\Output{$Arc$: promising archive. }
	\BlankLine
	$P\leftarrow$ initialize a population of size $N$;\; 
	$F\leftarrow$ evaluate $P$ on validation set based on $S_w$;\;
	$L_q\leftarrow $ mark the labels for $P$ on the basis of $F$;\;
	$t\leftarrow$ 0, $Arc\leftarrow P$, $Q\leftarrow \emptyset$;\;
	\While{$t<T$}{
		{\footnotesize{\tcp{Selection of reference points.}}}
		$S_R\leftarrow$ Select-Reference ($P$);\;
		{\footnotesize{\tcp{Classifier updating and validation.}}}
		$Sc\leftarrow$ Classification-Label ($S_R$, $Arc$);\;
		$[D_{train}, D_{val}]\leftarrow$ Data-Division ($Arc$, $S_c$); \;
		Classifier $\leftarrow$ Train-Ensemble ($D_{train}$); \;
		$AUC\leftarrow$ Validation (Classifier, $D_{val}$);\;
		{\footnotesize{\tcp{Classifier-assisted environmental selection.}}}
		$Q\leftarrow$ Reproduction ($P\cup S_R$);\;
		$P\leftarrow$ CESelection (Classifier, $Q$, $P$, $AUC$, $Th\_AUC$);\;
	    $Arc\leftarrow$ select 50\% from $P$ according to NSGA-II \cite{67r};\;
		$t \leftarrow t + 1$; \;
	}
	\textbf{Return} $Arc$;
	\caption{Main Framework of CENAS}\label{Algori1}
\end{algorithm}


\subsection{Search space encoding}\label{SSE}
Following \cite{12r, 16r}, the search space in this study is based on cells, i.e., normal and reduction cells. The normal cells preserve the input spatial size, while the reduction cells reduce the input spatial size. Aiming to find the suitable architectures inside the cells, we use densely-connected-like cells. 

\begin{figure}[htbp]
	\centering
	\includegraphics[width=\linewidth]{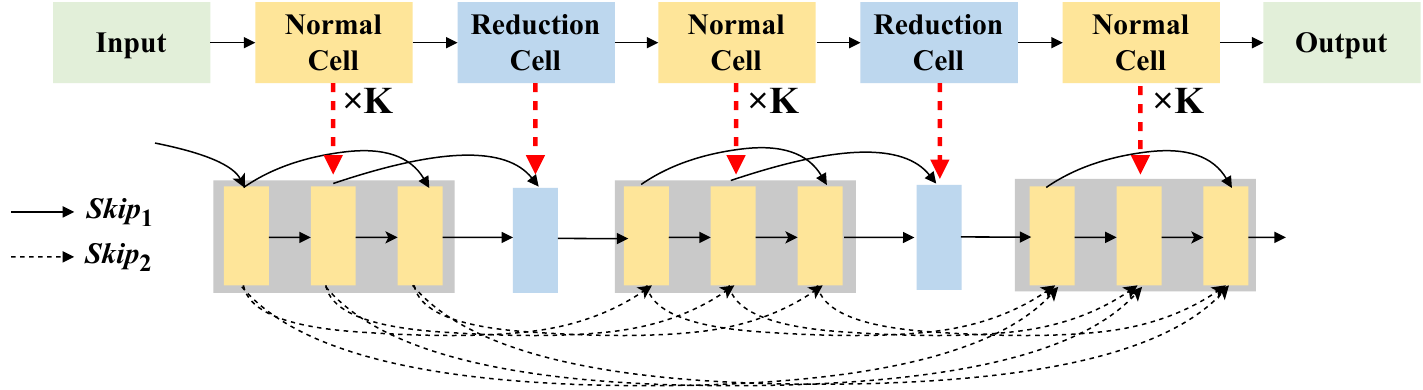}\\
	\caption{An illustration of the search space of CENAS. Top: the outer structure (omitting skip inputs for clarity) for architectural search. Bottom: an example to show the densely-connection via the skip operations when $K$ is set to three, where the normal cell group consists of three normal cells.}\label{Fig2a}
\end{figure}

The search space has a fixed outer structure as shown in Fig. \ref{Fig2a} (Top): stacking cells in a feed-forward way. An example of the densely-connection is shown in Fig. \ref{Fig2a} (Bottom). Each cell has a $direct$ $input$ \cite{23r} from the previous cell and some $skip$ $input$s \cite{23r} from the cells via $skip$ operations ($Skip\/_1$ or/and $Skip\/_2$). Specifically, $Skip\/_1$ is either to connect two normal cells in the normal cell group, or to link the normal cell and adjacent reduction cell. In contrast, $Skip\/_2$ is to connect two normal cells from two different normal cell groups via down-sampling operation (or a pooling operation). 

\begin{figure}[t]
	\centering
	\includegraphics[width=\linewidth]{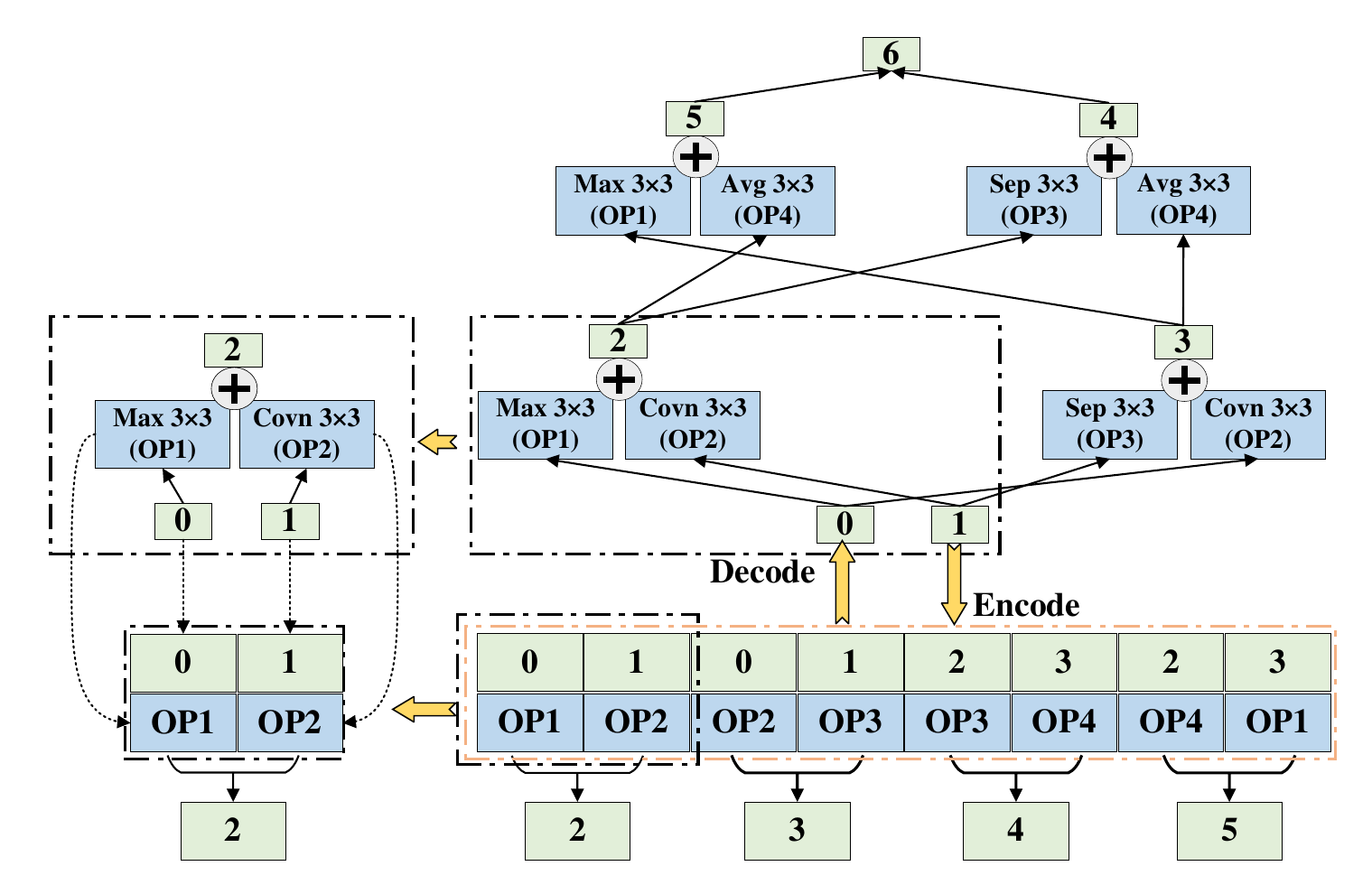}\\
	\caption{Left: An example of a pairwise combination and corresponding encoding. Top: an example to illustrate the structure of a cell represented by a directed acyclic graph with seven nodes highlighted in green (nodes 0,$\ldots$,6). Bottom: an example of encoding matrix to represent the above cell.}\label{Fig2b}
\end{figure}

For each cell, it has seven hidden states (or nodes), i.e., two input, four intermediate, and one output hidden state. Specifically, the input hidden states are the \textit{direct input} and \textit{skip input}, and output hidden state can receive the output of any of the hidden states. Very first cell takes two copies of the direct input as the input hidden states for consistency. An example of the structure of a cell is shown in Fig. \ref{Fig2b} (Top) with seven hidden states (i.e., the green blocks). The input hidden states of the cell are 0 and 1, while the output hidden state is 6. The intermediate hidden states 2-5 between the input and output hidden states all have two operations (i.e., blue blocks).

During the encoding of a cell, except the input hidden states (i.e., 0 and 1) and the output hidden state (i.e., 6) , all the hidden states (2-5) need to be encoded. Fig. 6 (Bottom) is an example of the encoding matrix. The first two columns of the matrix represent the hidden states 2, which can be decoded in Fig. \ref{Fig2b} (Left). Specifically, 0 and 1 mean the input hidden states. OP1 and OP2 are the operations of Max 3x3 and Conv 3x3, respectively. The hidden state 2 is formed by the previous hidden states and operations. Similarly, we can get other hidden states.

		\begin{algorithm} 
			\DontPrintSemicolon
			\SetAlgoLined
			\SetKwInOut{Input}{Input}\SetKwInOut{Output}{Output}
			\Input{$P$: population.}
			\Output{$S_R$: reference points. }
			\BlankLine
			$\{PF_1, PF_2, \cdots\}\leftarrow$ Non-dominating sorting ($P$);\;
			$C\leftarrow$ Angle-based-clustering ($PF_1$);\;			 
			\While{$i < |C|$}{
			$\overline{c_i}\leftarrow$ the center of cluster $C_i$ is computed by Eq. (2);\;
			\eIf{$|C_i|\leq 3$}{
			{\footnotesize{\tcp{the center is selected as the reference point.}}}
				$S_R\leftarrow \overline{c_i}$; \;
			}{
			$B_i\leftarrow$ two solutions with the maximum acute angle in $C_i$;\;
			$S_R \leftarrow S_R \cup \overline{c_i}$;\;
			$S_R \leftarrow S_R \cup B_i$;\;
			} 
			}
			\textbf{Return} $S_R$;
			\caption{Select-Reference ($P, F$)}\label{Algori2}
		\end{algorithm}


\subsection{Selection of reference points}
\begin{figure}[t]
		\centering
		\begin{tabular}{@{}c@{}c@{}}
			\includegraphics[width=.45\linewidth]{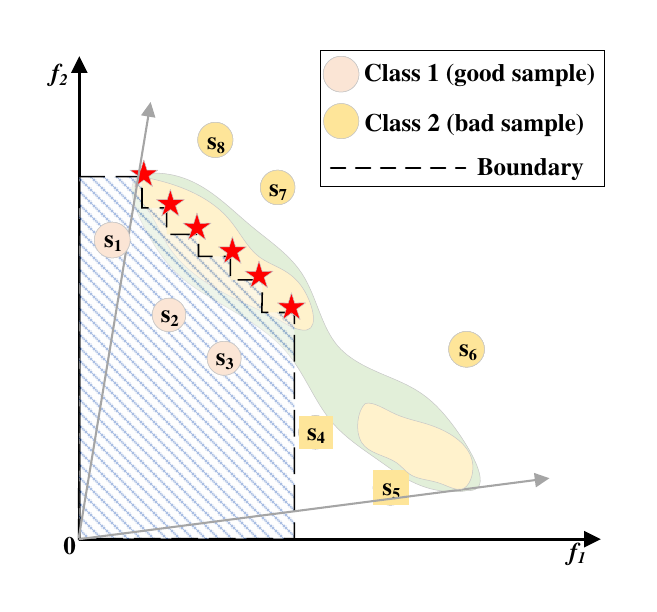}&\includegraphics[width=.45\linewidth]{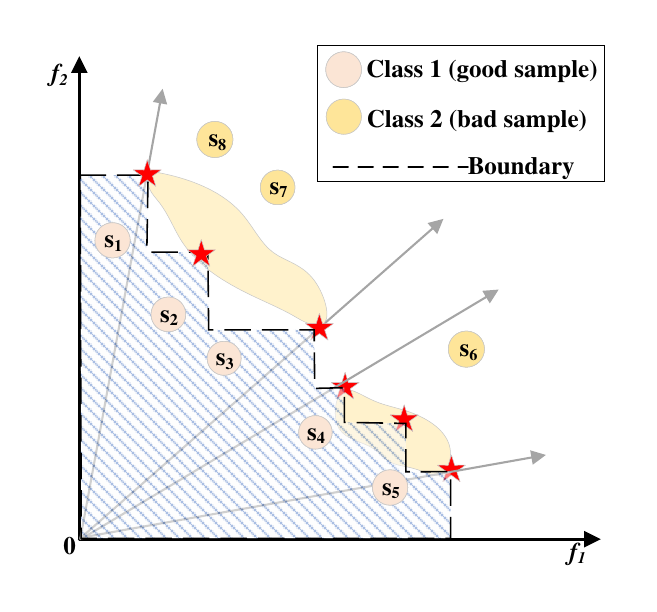}\\
			(a) & (b)
		\end{tabular}
\caption{An example to show the impact of different distributions of reference points (denoted by red stars) to the classification. The boundary of plot (a) is formed by six randomly selected reference points, while the boundary of plot (b) is formed by six different reference points from two segments of the front.}\label{Fig3}
\end{figure}

In order to build a robust ranking classifier to detect the candidate architectures whether they are ``good'' or not from different regions of the Pareto front, we have to appropriately choose a number of well-distributed reference architectures as the classification boundary. An example is given in Fig. \ref{Fig3}. On the basis of the classification boundary in plot \ref{Fig3} (a), $s_1$--$s_3$ belong to the good class, while the rest to the poor class. Apparently, the trained classifier may be biased and the architectures such as $s_4$ and $s_5$ are misclassified. In contrast, the classification based on the boundary in plot \ref{Fig3} (b) shows better performance, which is able to reflect the distribution of the architectures.

According to the above analysis, we have designed an adaptive clustering-based approach to get set of well-distributed representative reference points as the classification boundary. The pseudocode of the approach is given in Algorithm \ref{Algori2}, and the main steps are described as follows:

\subsubsection{Non-domination sorting (Line 1 of Algorithm \ref{Algori2})} A number of non-dominated fronts ($PF_1$, $PF_2$, $\cdots$) are obtained via the conventional non-dominated sorting \cite{67r} on the population from scratch \cite{18r}, where the solutions in each front are non-dominated with each other. The non-dominated sorting is introduced in Section I of the Supplementary material.

\subsubsection{Angle-based clustering (Line 2 of Algorithm \ref{Algori2})} In this step, the first front ($PF_1$) is grouped into a set of clusters ($C$) by the proposed angle-based-clustering. Here, the proposed clustering aims to find solutions with high similarities in the first front into the same cluster, and the clustering is based on the hierarchical clustering algorithm (HCA) \cite{36r}. Notably, we use the angle-based distances between intra-class samples (in Eq. \ref{Eq2}) rather than the Euclidean distances for the evaluation of the similarity between the samples. The advantage is that there is no need to preset the number of clusters and hold the assumption of a normal distribution of data. More details of HCA are given in Section II of the Supplementary material.

The similarity between two architectures can be measured by their acute angle:
\begin{equation}\label{Eq2}
    angle(X,Y)=|\frac{\sum_{i=1}^{m}\left(X_{i} \times Y_{i}\right)}{\sqrt{\sum_{i=1}^{m}\left(X_{i}\right)^{2}} \times \sqrt{\sum_{i=1}^{m}\left(Y_{i}\right)^{2}}}|,
\end{equation}
\noindent where $m$ is the number of objectives. $X$ and $Y$ are two vectors of objective values, and $X_i$ and $Y_i$ denote their $i$th objective values, respectively.

Following the above clustering, we can obtain a cluster set $C=\{C_1,C_2,\ldots,C_k\}$, where $k$ is the number of clusters and adaptively determined by the clustering.

\subsubsection{Selection (Lines 3-12 of Algorithm \ref{Algori2})} After the clustering, we can get the center and boundary points of each cluster, in order to acquire a number of reference points from these clusters. The details are as follows: 

After objective normalization, we compute the cluster center by averaging the objective values of the solutions in the corresponding cluster:
\begin{equation}\label{Eq3}
	\overline{c_i}= (\frac{\sum_{j=1}^{|C_i|}f_1(x_j)}{|C_i|},\ldots,\frac{\sum_{j=1}^{|C_i|}f_m(x_j)}{|C_i|}) ,		
\end{equation}
where $\overline{c_i}$ is the center of cluster $C_i$.

Once the cluster centers are determined, we can calculate the acute angle between each pair of solutions in the cluster. Then, the two solutions with the maximum acute angle are selected as the boundary points of the cluster. 

Based on the number ($|C_i|$) of solutions in a cluster $C_i$, we have two cases for the selection of reference points from the cluster where $N_c$ is a threshold:
\begin{itemize}
\item {\textbf{Case 1}}: If $|C_i| \leq N_c$, then add the cluster center into the set of reference points.
\item {\textbf{Case 2}}: If $|C_i| > N_c$, then add the center and two boundary solutions into the set of reference points. 
\end{itemize}

Following the above procedures, we can get a set of effective reference points.

\subsection{Classifier updating and validation}
To reduce the high computation cost of the subnet evaluations under multiple objectives, we construct an online Pareto ranking classifier to directly predict the Pareto-dominance relationship between a candidate subnet and reference points. Especially, a number of low-complexity basic SVMs \cite{38r} are chosen to build an ensemble of SVM (ESVM) as the classifier, since SVM does not require much training data and is not prone to incur overfitting. The pseudocode is presented in Algorithm \ref{Algori3} and the main steps are as follows.

\subsubsection{Sample labelling}
Once the reference points are acquired, we then label the samples for the training of the comparator, as shown in Algorithm \ref{Algori3}.

\textbf{Data labelling} (Lines 1-12 of Algorithm \ref{Algori3}): The traditional Pareto dominance comparison is used to label the samples in the following way: If a solution is Pareto dominated by any reference point, then it will be labeled as a negative sample; otherwise, it is positive.

\textbf{Class imbalance} (Lines 13-26 of Algorithm \ref{Algori3}): During the evolutionary process, the proportion of ``good'' and ``poor'' solutions is dynamically changing, where the number of ``poor'' solutions is larger than that of the ``good'' solutions in the early stage but tends to be smaller in the late stage of the optimization. Thus, the class imbalance phenomenon\footnote{The class imbalance means that the rate of positive data over negative data is unreasonable (e.g., <10\% or >90\%).} may occur in the labeling process, which may decrease the prediction performance of the comparator. To tackle this problem, we utilize an $\alpha$-domination criterion \cite{37r} to relabel the samples after clustering \footnote{Our experimental results are shown in Fig \ref{Figbr} in Section \ref{S-IV}, which indicate that the clustering and $\alpha$-domination play a positive role in handling the imbalance issue.}.

\textbf{Definition 1} ({\it{$\alpha$-domination criterion}}\cite{37r})
A solution $p$ is considered to $\alpha$-dominate another solution $q$ (denoted as $p \prec_{\alpha} q$), if the following conditions are satisfied:
\begin{equation}\label{Def1}
\begin{aligned}
 & \forall i \in \{1,2,\ldots, m\},  h_i(p,q) \leq 0 \quad \wedge \\
 & \exists j  \in \{1,2,\ldots, m\},  h_j(p,q) < 0,
\end{aligned}
\end{equation}
where $h_i(p,q)=f_i(p)-f_i(q)+\sum^{m}_{j\neq i}\alpha_{ij}(f_j(p)-f_j(q))$, and $\alpha_{ij}$ is the predefined bound of the trade-off rates. $f_i(p)$ is the $i$th objective function of solution $p$.

\begin{figure}[t]
		\centering
		\begin{tabular}{@{}c@{}c@{}}
			\includegraphics[width=.45\linewidth]{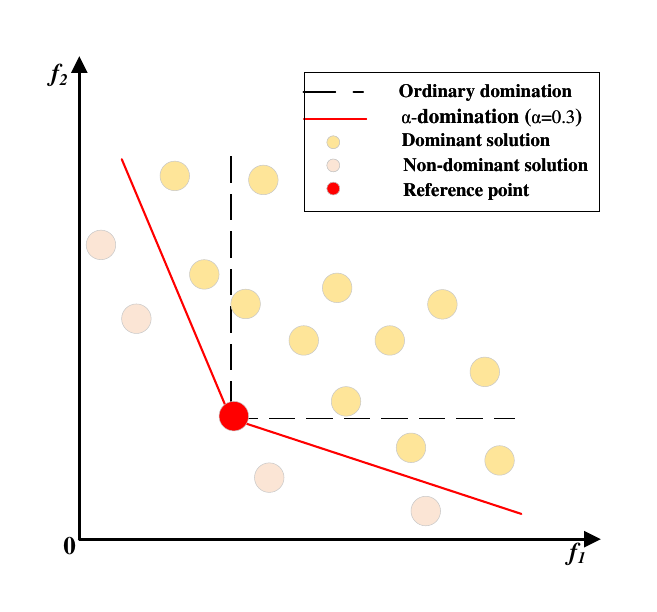}&\includegraphics[width=.45\linewidth]{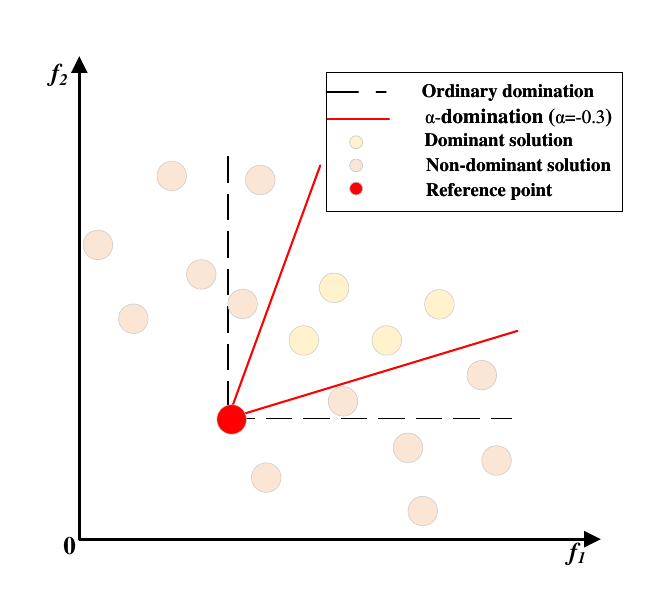}\\
			(a) & (b)
		\end{tabular}
\caption{An example to show the difference of the number of dominated solutions by using the $\alpha$-domination with different $\alpha$. (a) a larger $\alpha$ leads to a larger dominated region of a solution and smaller number of positive samples will be obtained. (b) a smaller $\alpha$ leads to smaller dominated region and more positive samples.}\label{Fig4}
\end{figure}

An example of the domination is shown in Fig. \ref{Fig4}. We can find that the $\alpha$-domination with different $\alpha$ values is able to control different sizes of dominated area. By this way, the proportion between two classes of samples can be regulated to an acceptable level, so that the class imbalance problem will be resolved. For example, if the negative samples are in high proportion, we may use a small $\alpha$ to decrease their numbers and increase the number of positive samples. 

To this end, an imbalance rate ($IR$) is defined in Eq. \ref{Eq5} to decide whether to activate the $\alpha$-domination criterion:
\begin{equation}\label{Eq5}
	IR= \frac{n_{majority}}{n_{minority}},		
\end{equation}
where $n_{majority}$ and $n_{minority}$ are the number of samples in the majority and minority categories, respectively. Here, the class imbalance occurs when $IR$ is larger than nine \cite{b1}.

\textbf{Data division} (Algorithm \ref{Algori4}): After the data labelling, the sample set ($Arc$) is divided into the training set ($D_{train}$) and validation set ($D_{val}$) for the training of ESVM. The rate of the size of $D_{train}$ over that of $D_{val}$ is empirically set to 3:2 in this work.

		\begin{algorithm} 
			\DontPrintSemicolon
			\SetAlgoLined
			\SetKwInOut{Input}{Input}\SetKwInOut{Output}{Output}
			\Input{$S_R$: set of reference points, $Arc$: archive.}
			\Output{$S_c$: categories of solutions. }
			\BlankLine
			$i\leftarrow 0$; \;	 
			\While{$i < |Arc|$}{
			$l\leftarrow$ False; {\footnotesize{\tcp{class 2: poor one.}}}
					\While{$ j < |SR|$}{
						 $l\leftarrow l \parallel$ ($S_{Rj}$ dominating $Arc_i$);\;
					}
					\eIf{ $l$ is False }{
 						$S_{c_i}\leftarrow$ False;\;
 					}{
						$S_{c_i}\leftarrow$ True; {\footnotesize{\tcp{class 1: good one.}}}
					}
			}
			$IR\leftarrow$ the imbalance rate is obtained by Eq. \ref{Eq5};\;
			\If{$IR>4$}{ 
			{\footnotesize{\tcp{the class distribution is imbalanced.}}}
					\While{$ i < |Arc| $}{
						 $l\leftarrow$ False;\;
						 \While{$  j < |SR| $}{
									  $l\leftarrow l \parallel $ ($S_{Rj}\prec_{\alpha} Arc_i$);\;
							}
							\eIf{$l$ is False }{
									$S_{c_i}\leftarrow$ False;\;
 								}{
									$S_{c_i}\leftarrow$ True; 
								}
					}

			}
			\textbf{Return} $S_c$;
			\caption{Classification-Label ($S_R$, $Arc$)}\label{Algori3}
		\end{algorithm}

\begin{algorithm} 
	\DontPrintSemicolon
	\SetAlgoLined
	\SetKwInOut{Input}{Input}\SetKwInOut{Output}{Output}
	\Input{$S_c$: categories of solutions, $Arc$: archive.}
	\Output{$[D_{train}, D_{val}]$: $[$Training set, Validation set$]$. }
	\BlankLine
	$D_1 \leftarrow$ find a solution with the label True in $Arc$; \;
	$D_2 \leftarrow$ find a solution with the label False in $Arc$;\;
	$d_1 \leftarrow$ randomly select 3/5 solutions from $D_1 \cup D_2$;\;
	$d_2 \leftarrow$ randomly select 2/5 solutions from $D_1 \cup D_2$;\;
	$D_{train}\leftarrow \{Arc_{d_1}, S_{c_{i1}}\}$;\;
	$D_{val} \leftarrow \{Arc_{d_2}, S_{c_{i2}}\}$;\;
	\textbf{Return} $[D_{train}, D_{val}]$;
	\caption{Data-Division ($S_c, Arc$)}\label{Algori4}
\end{algorithm}

\subsubsection{Classifier updating and validation}
The classifier updating and validation are shown in Lines 9-10 of Algorithm \ref{Algori1}. 

\textbf{Classifier updating} (Algorithm \ref{Algori5}): We adopt the ensemble SVM (ESVM)\footnote{Our experimental results (as shown in Fig. \ref{Fig9} and Fig \ref{Figc} in Section \ref{S-IV}) demonstrate that ESVM performs better and with much less samples than other surrogate models.} with Bagging \cite{38r} to classify the candidate architectures. The reason is that the ESVM has been verified to be able to reduce the probability of overfitting, and ensure the effectiveness of the comparator on a small dataset \cite{38r}. Specifically, each SVM is firstly independently trained using the bootstrap approach, which can tackle the issue of sample reduction caused by Cross-Validation. Then, the majority voting is employed as the aggregation rule to decide the final output results of the comparator on new data because it can decrease the error rate of comparator \cite{b2}.

\textbf{Model validation} (Line 10 in Algorithm \ref{Algori1}): In this step, a measurement ($AUC$\footnote{$AUC$ is referred to as the area under the curve of receiver operating characteristic curve \cite{39r}.}) \cite{39r} is adopted to evaluate the reliability of a classifier on the test data, where $AUC$ can be calculated by Eq. \ref{Eq6}. 

\begin{equation}\label{Eq6}
	AUC= \frac{\sum I(V_{Positive},V_{Negative})}{M*N},		
\end{equation}
where $M$ and $N$ represent the number of positive and negative samples in the data set, respectively. $V_{Positive}$ and $V_{Negative}$ are the predicted probability of positive and negative samples. $I(\cdot)$ is defined as 
\begin{equation}\label{Eq7}
	I(V_{Positive},V_{Negative}) = \left\{ \begin{array}{ll}
	1, & \textrm{$V_{Positive}>V_{Negative}$}\\
	0.5, & \textrm{$V_{Positive}= V_{Negative}$}\\
	0, & \textrm{$V_{Positive}< V_{Negative}$}
	\end{array} \right..
\end{equation}
In principle, a higher $AUC$ indicates a better prediction reliability of a classifier.

\begin{figure}[t]
\centering
\includegraphics[width=0.8\linewidth]{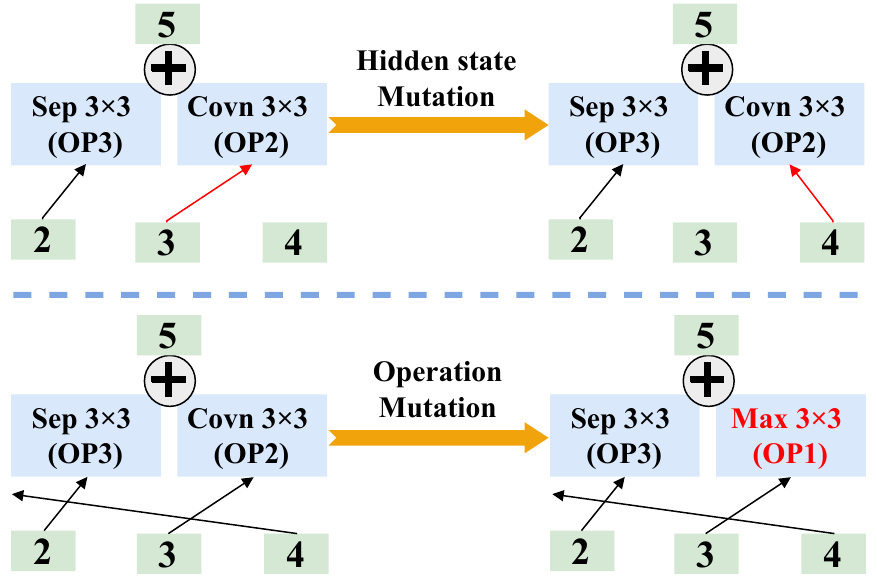}\\
\caption{Two mutation types in CENAS.}\label{Fig5}
\end{figure}

\begin{algorithm} 
	\DontPrintSemicolon
	\SetAlgoLined
	\SetKwInOut{Input}{Input}\SetKwInOut{Output}{Output}
	\Input{$D_{train}$: training dataset, $K_s$: ensemble size.}
	\Output{ESVM: a set of base SVM learners. }
	\BlankLine
	$i\leftarrow 0$, ESVM$\leftarrow \emptyset$; \;
	\While{$ i < K $}{
		$D_{train_i}\leftarrow$ bootstrap sample from dataset $D_{train}$;\;
		SVM $\leftarrow$ construct an  SVM from $D_{train_i}$;\;
		ESVM $\leftarrow$ ESVM $\cup$ SVM; \;
		$i\leftarrow i + 1$;
	}
	\textbf{Return} ESVM; 
	\caption{Train-Ensemble ($D_{train}$) }\label{Algori5}
\end{algorithm}

\subsection{Classifier-assisted environmental selection}
During the evolutionary cycle, reproduction is performed on the existing architectures (Line 12 of Algorithm \ref{Algori1}) to generate the offspring architectures. According to the encoding of the search space, we employ two mutation operators \cite{29r}, i.e., hidden state mutation and operation mutation, to generate new architectures. Specifically, the hidden state mutation aims to create new architecture by changing the input (i.e., hidden state) of the operation. For example, the input of the operation (OP2) is changed from the hidden state 3 to the hidden state 4 in Fig. \ref{Fig5} (Top). The operation mutation is designed to modify the architecture by changing the operation of the hidden state. Such the operation (Covn 3$\times$3) is changed to Max 3$\times$3 in Fig. \ref{Fig5} (Bottom).

Then, the classifier-assisted selection is used to select architectures from the offspring population, as shown in Algorithm \ref{Algori6}. If the $AUC$ of the classifier is not reliable (i.e., $AUC < Th\_AUC$), the accuracy of all offspring individuals is evaluated by the weight inheritance (Line 2), and their other objectives can be directly calculated (Line 3). Otherwise, only the offspring individuals labelled as ``good/True'' solutions will be evaluated in a similar way (Lines 8-9). Finally, we use NSGA-II \cite{67r} to select $N$ individuals from the offspring $Q$ (Line 12), where the nondominated sorting and standard crowding distance are adopted to improve the convergence and diversity of the population ($P$). In addition, we provide the principles of classifier design in Section IV of the Supplementary material.

		\begin{algorithm} 
			\DontPrintSemicolon
			\SetAlgoLined
			\SetKwInOut{Input}{Input}\SetKwInOut{Output}{Output}
			\Input {Classifier: SVM ensemble, $Q$: offspring population, $P$: old population, $AUC$: an indicate value, $Th\_AUC$:predefined $AUC$ threshold, $S_w$:supernet.}
			\Output{$P$: population for next generation. }
			\BlankLine
			\eIf{$AUC < Th\_AUC$}{	  	
				{\footnotesize{\tcp{$F^*$ indicates the accuracy.}}}
				$F^*\leftarrow$ evaluate $Q$ by sharing weights of $S_w$;\;
				{\footnotesize{\tcp{$F_o$ indicates other objectives.}}}
				$F_o\leftarrow$ calculate other objective functions;\;
				$Q\leftarrow P\cup Q$;\;
			}
			{
				$L_q\leftarrow$ Prediction (Classifier, $Q$);\;
				$Q\leftarrow$ select solutions from $Q$ with $L_q = True$;\;
				$F^*\leftarrow$ evaluate $Q$ by sharing weights of $S_w$;\;
				$F_o\leftarrow$ calculate other objective functions;\;
				$Q\leftarrow P\cup Q$;\;
			}
			$P\leftarrow$ select $N$ solutions from $Q$ using NSGA-II \cite{67r};\;
			\textbf{Return} $P$; 
			\caption{CESelection (Classifier, $Q$, $P$, $AUC$, $Th\_AUC$, $S_w$)}\label{Algori6}
		\end{algorithm}

\section{Experimental Results and discussions}\label{S-IV}
\subsection{Experimental settings}
We employ five widely used image classification datasets for empirical studies, including ImageNet \cite{40r}, CIFAR-10 \cite{41r}, CIFAR-100 \cite{41r}, MNIST \cite{42r} and Fashion MNIST \cite{43r}. Table S2 of the Supplementary material illustrates their settings,  where the size of samples varies from 50,000 to 60,000 images (600 to 6,000 images per class).

The state-of-the-art NAS and manual approaches are used as the baselines for comparison, including AtomNAS \cite{1r}, ResNet50 \cite{7r}, NSGANET \cite{8r}, GeNet \cite{10r}, NASNet-A \cite{14r}, DARTS \cite{16r}, MnasNet \cite{19r}, AmoebaNet-A \cite{23r}, PNAS-5 \cite{24r}, EffPNet-B0 \cite{26r}, ProxylessNAS-RL \cite{27r}, OFA-random \cite{28r}, MobileNetV1 \cite{44r}, MobileNetV2 \cite{45r}, ShuffleNetV1 \cite{46r}, ShuffleNetV2 \cite{47r}, Path-level EAS \cite{48r}, PC-DARTS \cite{49r}, DSNAS \cite{50r}, FBNet \cite{51r}, SemiNAS \cite{52r}, RENASNet \cite{53r}, and DPP-Net \cite{54r}. Their experimental data follows the original literature. The hyperparameters of CENAS are configured in Table S3 of the Supplementary material, which has been validated in the ablation studies in Section VIII of the Supplementary material. The experimental results on CIFAR-10, CIFAR-100, MNIST and Fashion MNIST can be found in Section VII of the Supplementary material. Following \cite{7r,28r}, each model is further fine-tuned with 60 epochs by SGD optimizer, where the weight decay is set to 3e-4, the momentum is 0.9, the batch size is 64 and the learning rate is 0.025. According to the regularization setting \cite{26r}, both drop connect ratio and dropout ratio are set to 0.2. CENAS is implemented based on Pytorch 1.9 and runs on a platform equipped with four identical GeForce RTX 2080Ti GPU cards.

\subsection{Results on ImageNet}

\begin{table*}
	\centering
	\caption{Comparison with manual and automated design baselines on ImageNet, where ``-'' denotes no results provided. Bayesian means Bayesian-based optimization; RS refers to Random search. Surrogate technique here refers to a function which is used to map the architecture and performance. Modular search space means to transform global search into a modular search strategy (e.g., cell-based search space), and incomplete training means to use acceleration strategy (e.g., weight sharing) instead of training the network from scratch. The listed results of the baselines come from their reported data.}\label{Tb2}
	\resizebox{\linewidth}{!}{
		\begin{tabular}{c|c|c|c|c|c|c|c|c|c}\hline
			\multirow{2}{*}{Model}  & \#Params & FLOPS & Search cost & \multirow{2}{*}{Top-1 (\%)}& \multirow{2}{*}{Top-5 (\%)} &\multirow{2}{*}{Search method}& \multicolumn{3}{c}{Optimization strategy}\\\cline{8-10}
			& (M)   & (M) & (GPU days) &     &     &    & Surrogate technique & Modular search space & Incomplete training\\ \hline
			ResNet50 \cite{7r}	&	25.6	&	4100	&	--	&	75.3	&	92.2	&	Manual	&	--	&	--	&	--	\\\hline
			MobileNetV1 \cite{44r}	&	4.2	&	575	&	--	&	70.6	&	89.5	&	Manual	&	--	&	--	&	--	\\\hline
			MobileNetV2 \cite{45r}	&	3.4	&	300	&	--	&	74.7	&	91	&	Manual	&	--	&	--	&	--	\\\hline
			ShuffleNetV1 \cite{46r}	&	3.4	&	292	&	--	&	71.5	&	89.8	&	Manual	&	--	&	--	&	--	\\\hline
			ShuffleNetV2 \cite{47r}	&	3.5	&	299	&	--	&	72.6	&	--	&	Manual	&	--	&	--	&	--	\\\hline
			ProxylessNAS-RL \cite{27r}	&	5.8	&	465	&	8.3	&	74.6	&	92.3	&	RL	&	\checkmark	&	\checkmark	&	\checkmark	\\\hline
			MnasNet \cite{19r}	&	5.2	&	312	&	1666	&	76.7	&	93.3	&	RL	&	--	&	\checkmark	&	--	\\\hline
			EffPNet-B0 \cite{26r}	&	5.3	&	390	&	--	&	77.3	&	93.5	&	RL	&	--	&	\checkmark	&	--	\\\hline
			Path-level EAS \cite{48r}	&	--	&	594	&	8.3	&	74.6	&	91.9	&	RL	&	--	&	\checkmark	&	\checkmark	\\\hline
			NASNet-A \cite{14r}	&	5.3	&	564	&	2000	&	74	&	91.6	&	RL	&	--	&	\checkmark	&	--	\\\hline
			DARTS \cite{16r}	&	4.7	&	574	&	4	&	73.3	&	81.3	&	Gradient	&	--	&	\checkmark	&	\checkmark	\\\hline
			AtomNAS \cite{1r}	&	5.9	&	363	&	--	&	77.6	&	93.6	&	Gradient	&	--	&	--	&	\checkmark	\\\hline
			PC-DARTS \cite{49r}	&	5.3	&	597	&	3.7	&	75.8	&	92.7	&	Gradient	&	--	&	\checkmark	&	\checkmark	\\\hline
			DSNAS \cite{50r}	&	--	&	324	&	17.5	&	77.4	&	91.5	&	Gradient	&	--	&	\checkmark	&	\checkmark	\\\hline
			FBNet \cite{51r}	&	5.5	&	375	&	9	&	74.9	&	--	&	Gradient	&	--	&	--	&	\checkmark	\\\hline
			OFA-random \cite{28r}	&	7.7	&	230	&	40	&	76.9	&	--	&	RS	&	\checkmark	&	--	&	\checkmark	\\\hline
			PNAS-5 \cite{24r}	&	5.1	&	588	&	225	&	74.2	&	92.7	&	Bayesian	&	\checkmark	&	\checkmark	&	\checkmark	\\\hline
			SemiNAS \cite{52r}	&	6.32	&	599	&	4	&	76.5	&	93.2	&	Bayesian	&	\checkmark	&	\checkmark	&	\checkmark	\\\hline
			GeNet \cite{10r}	&	156	&	--	&	17	&	72.1	&	90.2	&	Evolution	&	--	&	--	&	--	\\\hline
			NSGANET \cite{8r}	&	5	&	585	&	27	&	76.2	&	93	&	Evolution	&	\checkmark	&	\checkmark	&	--	\\\hline
			AmoebaNet-A \cite{23r}	&	6.4	&	555	&	3150	&	74.5	&	92.4	&	Evolution	&	--	&	\checkmark	&	--	\\\hline
			RENASNet \cite{53r}	&	5.36	&	580	&	6	&	75.7	&	92.6	&	Evolution	&	\checkmark	&	\checkmark	&	--	\\\hline
			DPP-Net \cite{54r}	&	4.8	&	532	&	2	&	74	&	91.8	&	Evolution	&	\checkmark	&	\checkmark	&	\checkmark	\\\hline
			CENAS-A	&	\textbf{3.24}	&	\textbf{276} 	&	\textbf{1.91}	&	77.4	&	92.8	&	Evolution	&	\checkmark	&	\checkmark	&	\checkmark	\\\hline
			CENAS-B	&	4.08	&	396	&	\textbf{1.91}	&	78.9	&	93.6	&	Evolution	&	\checkmark	&	\checkmark	&	\checkmark	\\\hline
			CENAS-C	&	4.98	&	482	&	\textbf{1.91}	&	\textbf{79.6}	&	\textbf{94.1}	&	Evolution	&	\checkmark	&	\checkmark	&	\checkmark	\\\hline
		\end{tabular}
	}
\end{table*}

In this test, the performance of CENAS will be validated on ImageNet, showing its ability in handling multiple objectives, i.e., minimizing  the number of parameters (\#Params) and floating point operations per second (FLOPs), and maximizing the accuracy. All results are obtained under the termination condition with 50 generations. After that, we select three models (CENAS-A, B, and C) from the final solutions (the range of their \#Params are from 2M to 5M, which is suitable to resource-constrained devices) via the approach of trade-off decision analysis (introduced in Section III of the Supplementary material).
\begin{figure}[htbp]
	\centering
	\includegraphics[width=0.8\linewidth]{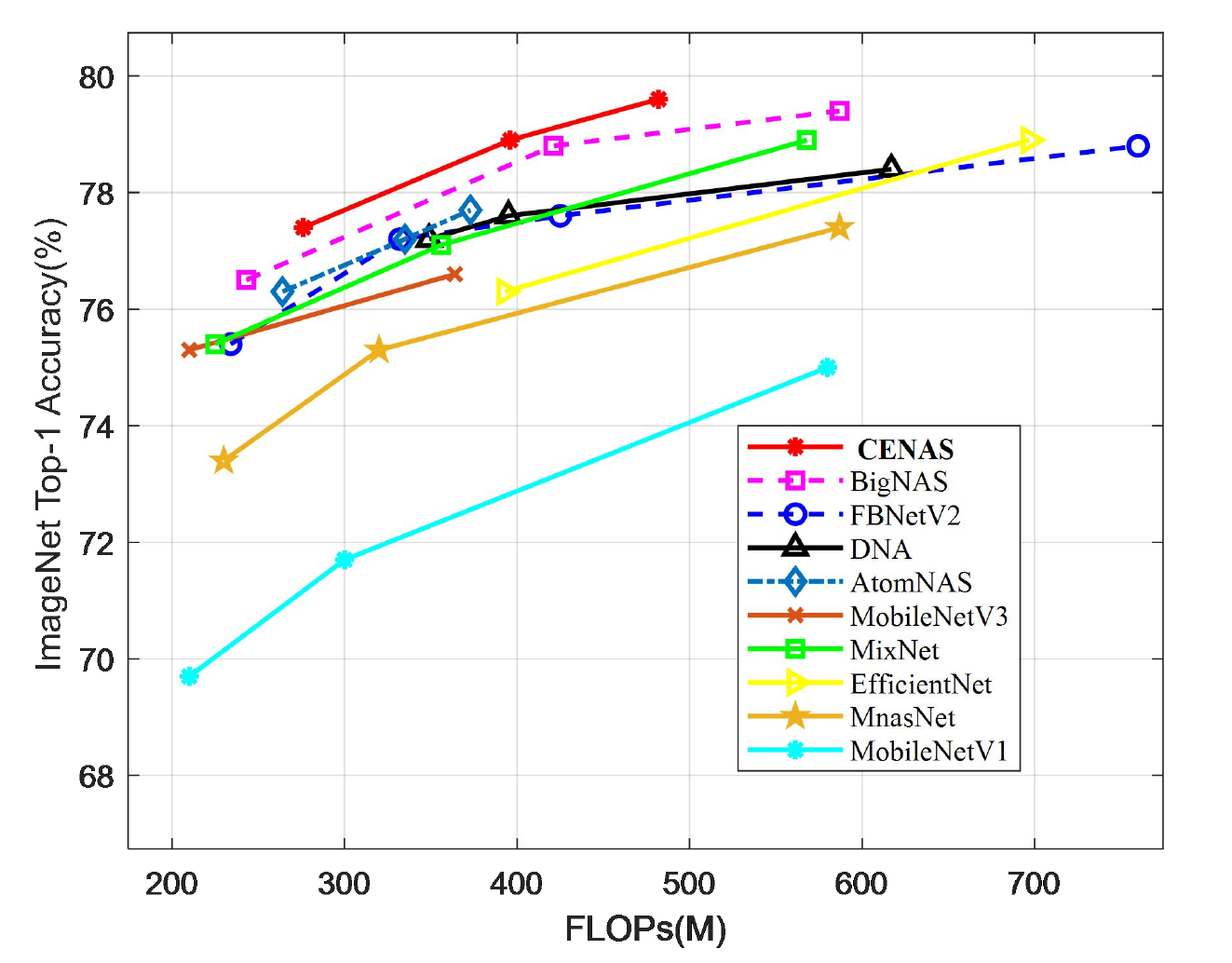}\\
	\caption{The trade-off fronts with respect to the FLOPs and Top-1 accuracy on ImageNet.}\label{Fig7}
\end{figure}

Table \ref{Tb2} reports the comparison results on ImageNet, in which \#Params, FLOPs, search cost and test accuracy (i.e., Top-1 and Top-5) obtained by all methods are listed. From Table \ref{Tb2}, we can see that CENAS-A performs the best with the smallest \#Params, and the second-best with the smallest FLOPs (only slightly worse than OFA-random, but the prediction accuracy of OFA-random with smaller FLOPs is much worse than that of CENAS-A). Especially, CENAS achieves higher accuracy than the manual design models (e.g., MobileNetV2 and ShuffleNetV2) with 4.9\% and 7\%, respectively. CENAS-A achieves smaller \#Params at the expense of a small degradation of accuracy, where the Top-5 accuracy of CENAS-A is slightly lower than that of GeNet and SemiNAS. CENAS-A also performs better than ResNet50, MobileNetV1 and ShuffleNetV1 in terms of \#Params. CENAS-C gets higher accuracy than RL-based methods including ProxylessNAS-RL, MnasNet, EfficientNet-B0 and Path-level EAS with 5\%, 2.9\%, 2.3\% and 5.6\%, respectively. Moreover, CENAS shows a significant search cost advantage over its competitors, consuming only 1.19 GPU days on ImageNet. In comparison with DPP-Net, our approach consumes 1.91 GPU days which is slightly smaller than that of DPP-Net with 2 GPU days. we find that the GPU days of DPP-Net does not include the labeling time of an architecture because it uses an offline surrogate model trained on a large amount of labeled architecture to satisfy the requirements for fitness evaluation during the search process. Note that the labeling of a neural network on CIFAR-10 takes 17 hours under normal configuration \cite{3r}. In contrast, our approach is based on an online model and we also count the labeling time in the GPU days. In addition, compared with other online surrogate model-assisted NAS (e.g., PNAS-5), CENAS achieves a larger improvement in the search efficiency. It may result from the fact that we integrate the ordinal optimization theory into our approach which can save much computational resources in discriminating the architectures. Especially, CENAS performs more powerfull than gradient-based NAS in both prediction accuracy and model size. In comparison with the surrogate-based NASs, CENAS dominates them (including ProxylessNAS-RL, NSGANET, AmoebaNet-A and DPP-Net) in terms of the performance indicator values. Therefore, the above results have initially validate the effectiveness of the proposed Pareto classifier.

In addition, Fig. \ref{Fig7} presents the Pareto fronts obtained by all methods with respect to the objectives (\#FLOPs and accuracy). The results indicate that CENAS has found the architecture with smaller \#FLOPs than most competitors. Although the \#FLOPs of CENAS are slightly larger than that of two competitors, the prediction accuracy of CENAS outperforms them with almost 4-7\%. CENAS achieves the accuracy with 79.6\% under the mobile setting ($\leq$500M FLOPs). Compared with BigNAS, the accuracy of CENAS is slightly higher than BigNAS when the value of FLOPs is around 400M while \#Params of CENAS is also smaller than BigNAS. Besides, the accuracy of CENAS is higher than that of MobileNetV3 \cite{60r} with 4.9\% under similar FLOPs.

There representative architectures (CENAS-A, B, and C) obtained by CENAS are visualized in Fig. S4 of the Supplementary material. Since CENAS-A has more skip$\_$connect and pooling operations compared with CENAS-B and CENAS-C, its \#Param is smaller than CENAS-B and CENAS-C. The accuracy of CENAS-A is not the highest either probably because of the lack of more feature extraction operations (e.g., sep$\_$conv and dil$\_$conv). 

Overall, CENAS shows strong competitive performance on ImageNet and achieves a significant efficiency improvement over its main competitors. The experimental results also show that CENAS achieves satisfactory performance in terms of both search efficiency and prediction accuracy. This may attribute to the efficiency of the classifier in handling the transformed binary Pareto-dominance classification task.

\subsection{Analysis of reference point selection and $\alpha$-domination} 
This experiment aims to validate the effectiveness of the selection of reference point and the influence of threshold $N_c$ on the learning performance of CENAS. Here, $N_c$ is set to 2, 3, 4, and 5 on CIFAR-10, CIFAR-100, MNIST, and Fashion MNIST, respectively. The results are shown in Fig. S8 in the Supplementary file. In terms of the results, we can observe that when $N_c$=3 CENAS achieves the best results on MNIST, Fashion MNIST and CIFAR-10, and the second best results on CIFAR-100. Especially, CENAS with $N_c$=3 performs better than other variants with different number of reference points after 150 epochs on CIFAR-100. It may lie in the fact that larger number of reference points may slow the speed of convergence, while smaller number of reference points may lead to poor diversity of the solutions. Therefore, the threshold $N_c$ is set to 3 in our experiments.

\begin{table}[htbp]
	\centering
	\caption{Four different strategies for the imbalance phenomenon}
	\begin{tabular}{c|c|c}
		\hline
		methods & Reference point selection & Domination strategy \\
		\hline
		A     & Clustering & $\alpha$-domination \\
		\hline
		B     & Clustering & Pareto domination \\
		\hline
		C     & Random & $\alpha$-domination \\
		\hline
		D     & Random & Pareto domination \\
		\hline
	\end{tabular}%
	\label{tab1}%
\end{table}%
\begin{figure}[htbp]
	\centering
	\includegraphics[width=0.8\linewidth]{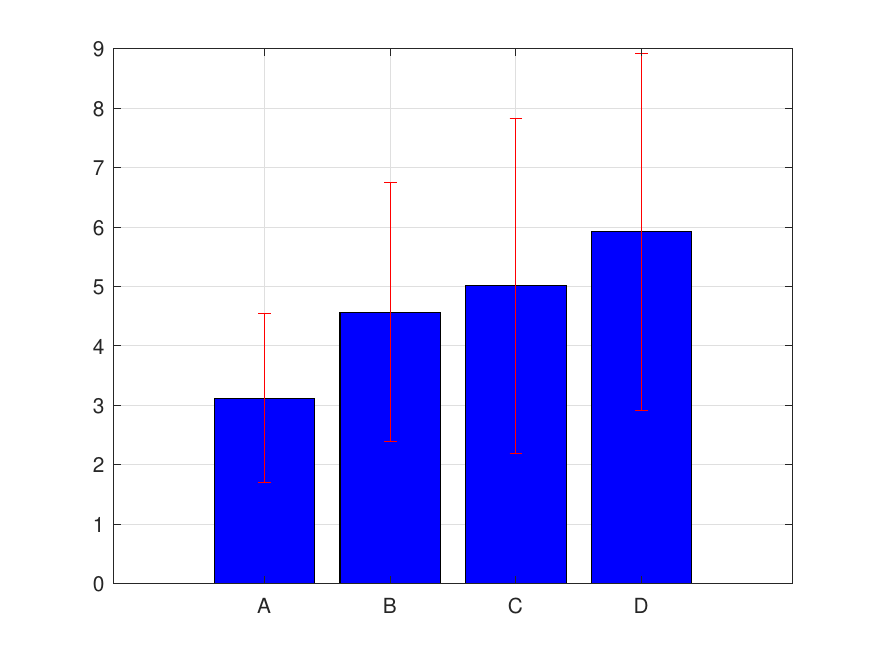}\\	
	\caption{The mean and variance of imbalance rates of different strategies on CIFAR-10.}\label{Figbr}	
\end{figure}

In order to study the advantages of the proposed method in solving the imbalance issue, we design four experimental scenarios based on different reference point selection and domination strategies, as shown in Table \ref{tab1}, where A is the proposed method. In this experiment, the imbalance rates between the ``majority '' and ``minority'' samples are recorded in all generations and Fig. \ref{Figbr} shows the means and variances of the imbalance rates of all strategies on CIFAR-10. We can observe that the mean and variance of A are much lower than that of other methods. Specifically, when A  is compared with B, we can find that $\alpha$-domination is more effective than the Pareto domination in handling the imbalance issue. Similar results can be achieved during the comparison of C and D. Another observation is that clustering is able to enhance the ability of  $\alpha$-domination to mitigate the imbalance issue when we compare A and C. Thus, the proposed method is able to provide with better classification boundary and more balanced training samples, which can solve the imbalance issue.

\subsection{Multi/Many-objective search ability}
We conduct another test to further validate the effectiveness and scalability of CENAS in dealing with complex NAS task with more objectives. At first, three objectives are taken into account, i.e., maximizing Top-1 accuracy, minimizing \#Params, and minimizing FLOPs. Then, six objectives, including the Top-1 accuracies on four datasets, \#Params and FLOPs, are optimized simultaneously. The corresponding experimental results and analysis are presented in Section IX of the Supplementary material due to the page limit. 

These results have shown that CENAS have achieved a trade-off front of multi-objective solutions with an appropriate balance between the convergence and diversity when multiple objectives are considered. In other words, CENAS has strong scalability in multi-objective NAS.

\section{Conclusion}\label{S-V}
In traditional neural architecture search, existing surrogates-driven methods commonly suffer from the issue of the ranking disorder and inefficiency in prediction. To alleviate the issue, we transformed the complex multi-objective NAS task into a simple classification task, and proposed a classifier-based Pareto evolution approach to handle the classification problem. Especially, an online end-to-end Pareto-wise ranking classifier is designed to predict the dominant relationship between the candidate and reference architectures. Besides, two adaptive schemes are designed for the reference point selection and mitigation of the class imbalance issue. A number of experiments are conducted, and the experimental results show that the proposed approach outperforms the state-of-the-art methods in efficiently finding a set of network architectures with different model sizes ranging from 2M to 5M under diverse objectives and constraints.

Despite the promising results, CENAS also encounters some threats: (1) From the internal aspect, there is a big margin for improvement in efficiently achieving a good balance between the convergence and diversity of the population through the appropriate selection of reference architectures; (2) From the external aspect, the challenge is how to largely reduce the highly computational expensive evaluations of the training samples on different datasets; (3) From the construction aspect, other effective prediction models for classification-based surrogates remain to be investigated.

Accordingly, there are some lines of future research. Firstly, we may take the knee points as the reference architectures to enhance the convergence and diversity of the population; Secondly, some training-free evaluation metrics can be adopted to evaluate the performance of the training samples, which can largely reduce the expensive evaluation cost \cite{a5}; Thirdly, we will investigate the effective end-to-end random forest model or listwise ranking model in our dominance prediction scenario.


\bibliographystyle{IEEEtran}
\bibliography{reference}

\begin{thebibliography}{10}
\providecommand{\url}[1]{#1}
\csname url@samestyle\endcsname
\providecommand{\newblock}{\relax}
\providecommand{\bibinfo}[2]{#2}
\providecommand{\BIBentrySTDinterwordspacing}{\spaceskip=0pt\relax}
\providecommand{\BIBentryALTinterwordstretchfactor}{4}
\providecommand{\BIBentryALTinterwordspacing}{\spaceskip=\fontdimen2\font plus
\BIBentryALTinterwordstretchfactor\fontdimen3\font minus
  \fontdimen4\font\relax}
\providecommand{\BIBforeignlanguage}[2]{{%
\expandafter\ifx\csname l@#1\endcsname\relax
\typeout{** WARNING: IEEEtran.bst: No hyphenation pattern has been}%
\typeout{** loaded for the language `#1'. Using the pattern for}%
\typeout{** the default language instead.}%
\else
\language=\csname l@#1\endcsname
\fi
#2}}
\providecommand{\BIBdecl}{\relax}
\BIBdecl

\bibitem{1r}
J.~Mei, Y.~Li, X.~Lian, X.~Jin, L.~Yang, A.~Yuille, and J.~Yang, ``Atomnas:
  Fine-grained end-to-end neural architecture search,'' in \emph{International
  Conference on Learning Representations}, 2020.

\bibitem{2r}
N.~Li, L.~Ma, G.~Yu, B.~Xue, M.~Zhang, and Y.~Jin, ``Survey on evolutionary
  deep learning: Principles, algorithms, applications and open issues,''
  \emph{ACM Computing Surveys}, 2023,
  doi:\href{https://doi.org/10.1145/3603704}{https://doi.org/10.1145/3603704}.

\bibitem{3r}
Y.~Sun, X.~Sun, Y.~Fang, G.~G. Yen, and Y.~Liu, ``A novel training protocol for
  performance predictors of evolutionary neural architecture search
  algorithms,'' \emph{IEEE Transactions on Evolutionary Computation}, vol.~25,
  no.~3, pp. 524--536, 2021.

\bibitem{4r}
A.~Brock, T.~Lim, J.~M. Ritchie, and N.~Weston, ``{SMASH}: one-shot model
  architecture search through hypernetworks,'' in \emph{International
  Conference on Learning Representations}, 2018.

\bibitem{13r}
B.~Baker, O.~Gupta, N.~Naik, and R.~Raskar, ``Designing neural network
  architectures using reinforcement learning,'' in \emph{International
  Conference on Learning Representations}, 2017.

\bibitem{23r}
E.~Real, A.~Aggarwal, Y.~Huang, and Q.~V. Le, ``Regularized evolution for image
  classifier architecture search,'' in \emph{Proceedings of the AAAI Conference
  on Artificial Intelligence}, vol.~33, no.~01, 2019, pp. 4780--4789.

\bibitem{24r}
C.~Liu, B.~Zoph, M.~Neumann, J.~Shlens, W.~Hua, L.-J. Li, L.~Fei-Fei,
  A.~Yuille, J.~Huang, and K.~Murphy, ``Progressive neural architecture
  search,'' in \emph{Proceedings of the European Conference on Computer
  Vision}, 2018, pp. 19--34.

\bibitem{2rr}
N.~Li, L.~Ma, T.~Xing, G.~Yu, C.~Wang, Y.~Wen, S.~Cheng, and S.~Gao,
  ``Automatic design of machine learning via evolutionary computation: A
  survey,'' \emph{Applied Soft Computing}, p. 110412, 2023.

\bibitem{25r}
X.~Dai, P.~Zhang, B.~Wu, H.~Yin, F.~Sun, Y.~Wang, M.~Dukhan, Y.~Hu, Y.~Wu,
  Y.~Jia \emph{et~al.}, ``Chamnet: Towards efficient network design through
  platform-aware model adaptation,'' in \emph{Proceedings of the IEEE/CVF
  Conference on Computer Vision and Pattern Recognition}, 2019, pp.
  11\,398--11\,407.

\bibitem{26r}
M.~Tan and Q.~Le, ``Efficientnet: Rethinking model scaling for convolutional
  neural networks,'' in \emph{Proceedings of the International Conference on
  Machine Learning}.\hskip 1em plus 0.5em minus 0.4em\relax PMLR, 2019, pp.
  6105--6114.

\bibitem{14r}
B.~Zoph, V.~Vasudevan, J.~Shlens, and Q.~V. Le, ``Learning transferable
  architectures for scalable image recognition,'' in \emph{Proceedings of the
  IEEE/CVF Conference on Computer Vision and Pattern Recognition}, 2018, pp.
  8697--8710.

\bibitem{27r}
H.~Cai, L.~Zhu, and S.~Han, ``Proxylessnas: Direct neural architecture search
  on target task and hardware,'' in \emph{International Conference on Learning
  Representations}, 2019.

\bibitem{5r}
H.~Pham, M.~Guan, B.~Zoph, Q.~Le, and J.~Dean, ``Efficient neural architecture
  search via parameters sharing,'' in \emph{Proceedings of the International
  Conference on Machine Learning}.\hskip 1em plus 0.5em minus 0.4em\relax PMLR,
  2018, pp. 4095--4104.

\bibitem{6r}
G.~Bender, P.-J. Kindermans, B.~Zoph, V.~Vasudevan, and Q.~Le, ``Understanding
  and simplifying one-shot architecture search,'' in \emph{Proceedings of the
  International Conference on Machine Learning}.\hskip 1em plus 0.5em minus
  0.4em\relax PMLR, 2018, pp. 550--559.

\bibitem{16r}
H.~Liu, K.~Simonyan, and Y.~Yang, ``\protect{DARTS}: Differentiable
  architecture search,'' in \emph{International Conference on Learning
  Representations}, 2019.

\bibitem{28r}
H.~Cai, C.~Gan, T.~Wang, Z.~Zhang, and S.~Han, ``Once for all: Train one
  network and specialize it for efficient deployment,'' in \emph{International
  Conference on Learning Representations}, 2020.

\bibitem{29r}
L.~Li and A.~Talwalkar, ``Random search and reproducibility for neural
  architecture search,'' in \emph{Uncertainty in Artificial
  Intelligence}.\hskip 1em plus 0.5em minus 0.4em\relax PMLR, 2020, pp.
  367--377.

\bibitem{7r}
Z.~Lu, G.~Sreekumar, E.~Goodman, W.~Banzhaf, K.~Deb, and V.~N. Boddeti,
  ``Neural architecture transfer,'' \emph{IEEE Transactions on Pattern Analysis
  and Machine Intelligence}, vol.~43, no.~9, pp. 2971--2989, 2021.

\bibitem{30r}
S.~Xie, A.~Kirillov, R.~Girshick, and K.~He, ``Exploring randomly wired neural
  networks for image recognition,'' in \emph{Proceedings of the IEEE
  International Conference on Computer Vision}, 2019, pp. 1284--1293.

\bibitem{a6}
N.~Sinha and K.-W. Chen, ``Evolving neural architecture using one shot model,''
  in \emph{Proceedings of the Genetic and Evolutionary Computation Conference},
  2021, pp. 910--918.

\bibitem{12r}
Z.~Yang, Y.~Wang, X.~Chen, B.~Shi, C.~Xu, C.~Xu, Q.~Tian, and C.~Xu,
  ``\protect{CARS}: Continuous evolution for efficient neural architecture
  search,'' in \emph{Proceedings of the IEEE/CVF Conference on Computer Vision
  and Pattern Recognition}, 2020, pp. 1829--1838.

\bibitem{19r}
M.~Tan, B.~Chen, R.~Pang, V.~Vasudevan, M.~Sandler, A.~Howard, and Q.~V. Le,
  ``Mnasnet: Platform-aware neural architecture search for mobile,'' in
  \emph{Proceedings of the IEEE/CVF Conference on Computer Vision and Pattern
  Recognition}, 2019, pp. 2820--2828.

\bibitem{31r}
Z.~Li, T.~Xi, J.~Deng, G.~Zhang, S.~Wen, and R.~He, ``{GP-NAS}: Gaussian
  process based neural architecture search,'' in \emph{Proceedings of the
  IEEE/CVF Conference on Computer Vision and Pattern Recognition}, 2020, pp.
  11\,933--11\,942.

\bibitem{32r}
L.~Ma, J.~Cui, and B.~Yang, ``Deep neural architecture search with deep graph
  bayesian optimization,'' in \emph{2019 IEEE International Conference on Web
  Intelligence (WI)}.\hskip 1em plus 0.5em minus 0.4em\relax IEEE, 2019, pp.
  500--507.

\bibitem{8r}
Z.~Lu, I.~Whalen, V.~Boddeti, Y.~Dhebar, K.~Deb, E.~Goodman, and W.~Banzhaf,
  ``\protect{NSGA-NET}: A multi-objective genetic algorithm for neural
  architecture search,'' in \emph{Proceedings of the Genetic and Evolutionary
  Computation Conference}, 2019, pp. 419--427.

\bibitem{10r}
L.~Xie and A.~Yuille, ``Genetic {CNN},'' in \emph{Proceedings of the IEEE
  International Conference on Computer Vision}, 2017, pp. 1379--1388.

\bibitem{11r}
H.~Liu, K.~Simonyan, O.~Vinyals, C.~Fernando, and K.~Kavukcuoglu,
  ``Hierarchical representations for efficient architecture search,'' in
  \emph{International Conference on Learning Representations}, 2018.

\bibitem{sun2020automatically}
Y.~Sun, B.~Xue, M.~Zhang, G.~G. Yen, and J.~Lv, ``Automatically designing cnn
  architectures using the genetic algorithm for image classification,''
  \emph{IEEE Transactions on Cybernetics}, vol.~50, no.~9, pp. 3840--3854,
  2020.

\bibitem{a2}
Z.~Lu, I.~Whalen, Y.~Dhebar, K.~Deb, E.~D. Goodman, W.~Banzhaf, and V.~N.
  Boddeti, ``Multiobjective evolutionary design of deep convolutional neural
  networks for image classification,'' \emph{IEEE Transactions on Evolutionary
  Computation}, vol.~25, no.~2, pp. 277--291, 2021.

\bibitem{a3}
L.~Dudziak, T.~Chau, M.~Abdelfattah, R.~Lee, H.~Kim, and N.~Lane,
  ``{BRP}-{NAS}: Prediction-based nas using {GCNs},'' \emph{Advances in Neural
  Information Processing Systems}, vol.~33, pp. 10\,480--10\,490, 2020.

\bibitem{49r}
Y.~Xu, L.~Xie, X.~Zhang, X.~Chen, G.-J. Qi, Q.~Tian, and H.~Xiong,
  ``{PC-DARTS}: Partial channel connections for memory-efficient architecture
  search,'' in \emph{International Conference on Learning Representations},
  2020.

\bibitem{9r}
Y.-H. Kim, B.~Reddy, S.~Yun, and C.~Seo, ``\protect{NEMO}: Neuro-evolution with
  multiobjective optimization of deep neural network for speed and accuracy,''
  in \emph{Proceedings of the International Conference on Machine
  Learning}.\hskip 1em plus 0.5em minus 0.4em\relax PMLR, 2017, pp. 1--8.

\bibitem{sun2018igd}
Y.~Sun, G.~G. Yen, and Z.~Yi, ``Igd indicator-based evolutionary algorithm for
  many-objective optimization problems,'' \emph{IEEE Transactions on
  Evolutionary Computation}, vol.~23, no.~2, pp. 173--187, 2019.

\bibitem{15r}
Z.~Zhong, J.~Yan, W.~Wu, J.~Shao, and C.-L. Liu, ``Practical block-wise neural
  network architecture generation,'' in \emph{Proceedings of the IEEE/CVF
  Conference on Computer Vision and Pattern Recognition}, 2018, pp. 2423--2432.

\bibitem{a7}
Y.~Peng, A.~Song, V.~Ciesielski, H.~M. Fayek, and X.~Chang, ``{PRE}-{NAS}:
  predictor-assisted evolutionary neural architecture search,'' in
  \emph{Proceedings of the Genetic and Evolutionary Computation Conference},
  2022, pp. 1066--1074.

\bibitem{a8}
Y.~Peng, A.~Song, V.~Ciesielski, Fayek, and H.~M., ``{PRE}-{NAS}: Evolutionary
  neural architecture search with predictor,'' \emph{IEEE Transactions on
  Evolutionary Computation}, vol.~27, no.~1, pp. 26--36, 2023.

\bibitem{17r}
M.~Zhang, H.~Li, S.~Pan, X.~Chang, C.~Zhou, Z.~Ge, and S.~Su, ``One-shot neural
  architecture search: Maximising diversity to overcome catastrophic
  forgetting,'' \emph{IEEE Transactions on Pattern Analysis and Machine
  Intelligence}, vol.~43, no.~9, pp. 2921--2935, 2020.

\bibitem{21r}
L.~Pan, C.~He, Y.~Tian, H.~Wang, X.~Zhang, and Y.~Jin, ``A classification-based
  surrogate-assisted evolutionary algorithm for expensive many-objective
  optimization,'' \emph{IEEE Transactions on Evolutionary Computation},
  vol.~23, no.~1, pp. 74--88, 2018.

\bibitem{22r}
Y.-C. Ho, Q.-C. Zhao, and Q.-S. Jia, \emph{Ordinal optimization: Soft
  optimization for hard problems}.\hskip 1em plus 0.5em minus 0.4em\relax
  Springer Science \& Business Media, 2008.

\bibitem{a4}
Y.~Chen, Y.~Guo, Q.~Chen, M.~Li, W.~Zeng, Y.~Wang, and M.~Tan, ``Contrastive
  neural architecture search with neural architecture comparators,'' in
  \emph{Proceedings of the IEEE/CVF Conference on Computer Vision and Pattern
  Recognition}, 2021, pp. 9502--9511.

\bibitem{a1}
H.-G. Huang and Y.-J. Gong, ``Contrastive learning: An alternative surrogate
  for offline data-driven evolutionary computation,'' \emph{IEEE Transactions
  on Evolutionary Computation}, vol.~27, no.~2, pp. 370--384, 2023.

\bibitem{67r}
K.~Deb, ``A fast elitist non-dominated sorting genetic algorithm for
  multi-objective optimization: {NSGA-2},'' \emph{IEEE Transactions on
  Evolutionary Computation}, vol.~6, no.~2, pp. 182--197, 2002.

\bibitem{18r}
K.~Deb and H.~Jain, ``An evolutionary many-objective optimization algorithm
  using reference-point-based nondominated sorting approach, part {I}: solving
  problems with box constraints,'' \emph{IEEE Transactions on Evolutionary
  Computation}, vol.~18, no.~4, pp. 577--601, 2013.

\bibitem{36r}
Y.~Hua, Y.~Jin, and K.~Hao, ``A clustering-based adaptive evolutionary
  algorithm for multiobjective optimization with irregular {P}areto fronts,''
  \emph{IEEE Transactions on Cybernetics}, vol.~49, no.~7, pp. 2758--2770,
  2018.

\bibitem{38r}
F.~Mordelet and J.-P. Vert, ``A bagging {SVM} to learn from positive and
  unlabeled examples,'' \emph{Pattern Recognition Letters}, vol.~37, pp.
  201--209, 2014.

\bibitem{37r}
K.~Ikeda, H.~Kita, and S.~Kobayashi, ``Failure of {P}areto-based {MOEA}s: Does
  non-dominated really mean near to optimal?'' in \emph{Proceedings of the
  Congress on Evolutionary Computation}.\hskip 1em plus 0.5em minus 0.4em\relax
  IEEE, 2001, pp. 957--962.

\bibitem{b1}
R.~M. Mathew and R.~Gunasundari, ``A review on handling multiclass imbalanced
  data classification in education domain,'' \emph{2021 International
  Conference on Advance Computing and Innovative Technologies in Engineering
  (ICACITE)}, pp. 752--755, 2021.

\bibitem{b2}
H.~M. Gomes, J.~P. Barddal, F.~Enembreck, and A.~Bifet, ``A survey on ensemble
  learning for data stream classification,'' \emph{ACM Computing Surveys},
  vol.~50, no.~2, pp. 1--36, 2017.

\bibitem{39r}
A.~P. Bradley, ``The use of the area under the {ROC} curve in the evaluation of
  machine learning algorithms,'' \emph{Pattern Recognition}, vol.~30, no.~7,
  pp. 1145--1159, 1997.

\bibitem{40r}
O.~Russakovsky, J.~Deng, H.~Su, J.~Krause, S.~Satheesh, S.~Ma, Z.~Huang,
  A.~Karpathy, A.~Khosla, M.~Bernstein \emph{et~al.}, ``Imagenet large scale
  visual recognition challenge,'' \emph{International Journal of Computer
  Vision}, vol. 115, no.~3, pp. 211--252, 2015.

\bibitem{41r}
A.~Krizhevsky, G.~Hinton \emph{et~al.}, ``Learning multiple layers of features
  from tiny images,'' 2009.

\bibitem{42r}
H.~Larochelle, D.~Erhan, A.~Courville, J.~Bergstra, and Y.~Bengio, ``An
  empirical evaluation of deep architectures on problems with many factors of
  variation,'' in \emph{Proceedings of the International Conference on Machine
  Learning}, 2007, pp. 473--480.

\bibitem{43r}
H.~Xiao, K.~Rasul, and R.~Vollgraf, ``Fashion-mnist: a novel image dataset for
  benchmarking machine learning algorithms,'' \emph{arXiv preprint
  arXiv:1708.07747}, 2017.

\bibitem{44r}
A.~G. Howard, M.~Zhu, B.~Chen, D.~Kalenichenko, W.~Wang, T.~Weyand,
  M.~Andreetto, and H.~Adam, ``Mobilenets: Efficient convolutional neural
  networks for mobile vision applications,'' \emph{arXiv preprint
  arXiv:1704.04861}, 2017.

\bibitem{45r}
M.~Sandler, A.~Howard, M.~Zhu, A.~Zhmoginov, and L.-C. Chen, ``Mobilenetv2:
  Inverted residuals and linear bottlenecks,'' in \emph{Proceedings of the IEEE
  Conference on Computer Vision and Pattern Recognition}, 2018, pp. 4510--4520.

\bibitem{46r}
X.~Zhang, X.~Zhou, M.~Lin, and J.~Sun, ``Shufflenet: An extremely efficient
  convolutional neural network for mobile devices,'' in \emph{Proceedings of
  the IEEE Conference on Computer Vision and Pattern Recognition}, 2018, pp.
  6848--6856.

\bibitem{47r}
N.~Ma, X.~Zhang, H.-T. Zheng, and J.~Sun, ``Shufflenet v2: Practical guidelines
  for efficient cnn architecture design,'' in \emph{Proceedings of the European
  Conference on Computer Vision}, 2018, pp. 116--131.

\bibitem{48r}
H.~Cai, J.~Yang, W.~Zhang, S.~Han, and Y.~Yu, ``Path-level network
  transformation for efficient architecture search,'' in \emph{Proceedings of
  the International Conference on Machine Learning}.\hskip 1em plus 0.5em minus
  0.4em\relax PMLR, 2018, pp. 678--687.

\bibitem{50r}
S.~Hu, S.~Xie, H.~Zheng, C.~Liu, J.~Shi, X.~Liu, and D.~Lin, ``{DSNAS}: Direct
  neural architecture search without parameter retraining,'' in
  \emph{Proceedings of the IEEE/CVF Conference on Computer Vision and Pattern
  Recognition}, 2020, pp. 12\,084--12\,092.

\bibitem{51r}
B.~Wu, X.~Dai, P.~Zhang, Y.~Wang, F.~Sun, Y.~Wu, Y.~Tian, P.~Vajda, Y.~Jia, and
  K.~Keutzer, ``Fbnet: Hardware-aware efficient convnet design via
  differentiable neural architecture search,'' in \emph{Proceedings of the
  IEEE/CVF Conference on Computer Vision and Pattern Recognition}, 2019, pp.
  10\,734--10\,742.

\bibitem{52r}
R.~Luo, X.~Tan, R.~Wang, T.~Qin, E.~Chen, and T.-Y. Liu, ``Semi-supervised
  neural architecture search,'' \emph{Advances in Neural Information Processing
  Systems}, vol.~33, pp. 10\,547--10\,557, 2020.

\bibitem{53r}
Y.~Chen, G.~Meng, Q.~Zhang, S.~Xiang, C.~Huang, L.~Mu, and X.~Wang,
  ``Reinforced evolutionary neural architecture search,'' \emph{arXiv preprint
  arXiv:1808.00193}, 2018.

\bibitem{54r}
J.-D. Dong, A.-C. Cheng, D.-C. Juan, W.~Wei, and M.~Sun, ``Dpp-net:
  Device-aware progressive search for {P}areto-optimal neural architectures,''
  in \emph{Proceedings of the European Conference on Computer Vision}, 2018,
  pp. 517--531.

\bibitem{60r}
A.~Howard, M.~Sandler, G.~Chu, L.-C. Chen, B.~Chen, M.~Tan, W.~Wang, Y.~Zhu,
  R.~Pang, V.~Vasudevan \emph{et~al.}, ``Searching for mobilenetv3,'' in
  \emph{Proceedings of the International Conference on Computer Vision}, 2019,
  pp. 1314--1324.

\bibitem{a5}
W.~Chen, X.~Gong, and Z.~Wang, ``Neural architecture search on imagenet in four
  {GPU} hours: a theoretically inspired perspective,'' in \emph{International
  Conference on Learning Representations}.

\end{thebibliography}

\par\noindent 
\parbox[t]{\linewidth}{
	\noindent\parpic{\includegraphics[height=1in,width=0.66in,clip,keepaspectratio]{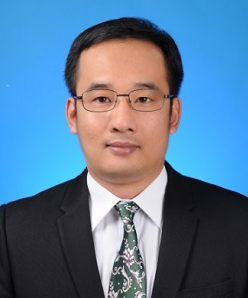}}
	\noindent {\bf Lianbo Ma}\ (Senior Member, IEEE) received the Ph.D. degree from the University of Chinese Academy of Sciences, Beijing, China, in 2015. He is currently a Professor with Northeastern University. He has published over 90 journal articles, books, and refereed conference papers. His current research interests include computational intelligence and machine learning.}

\par\noindent 
\parbox[t]{\linewidth}{
	\noindent\parpic{\includegraphics[height=1in,width=0.66in,clip,keepaspectratio]{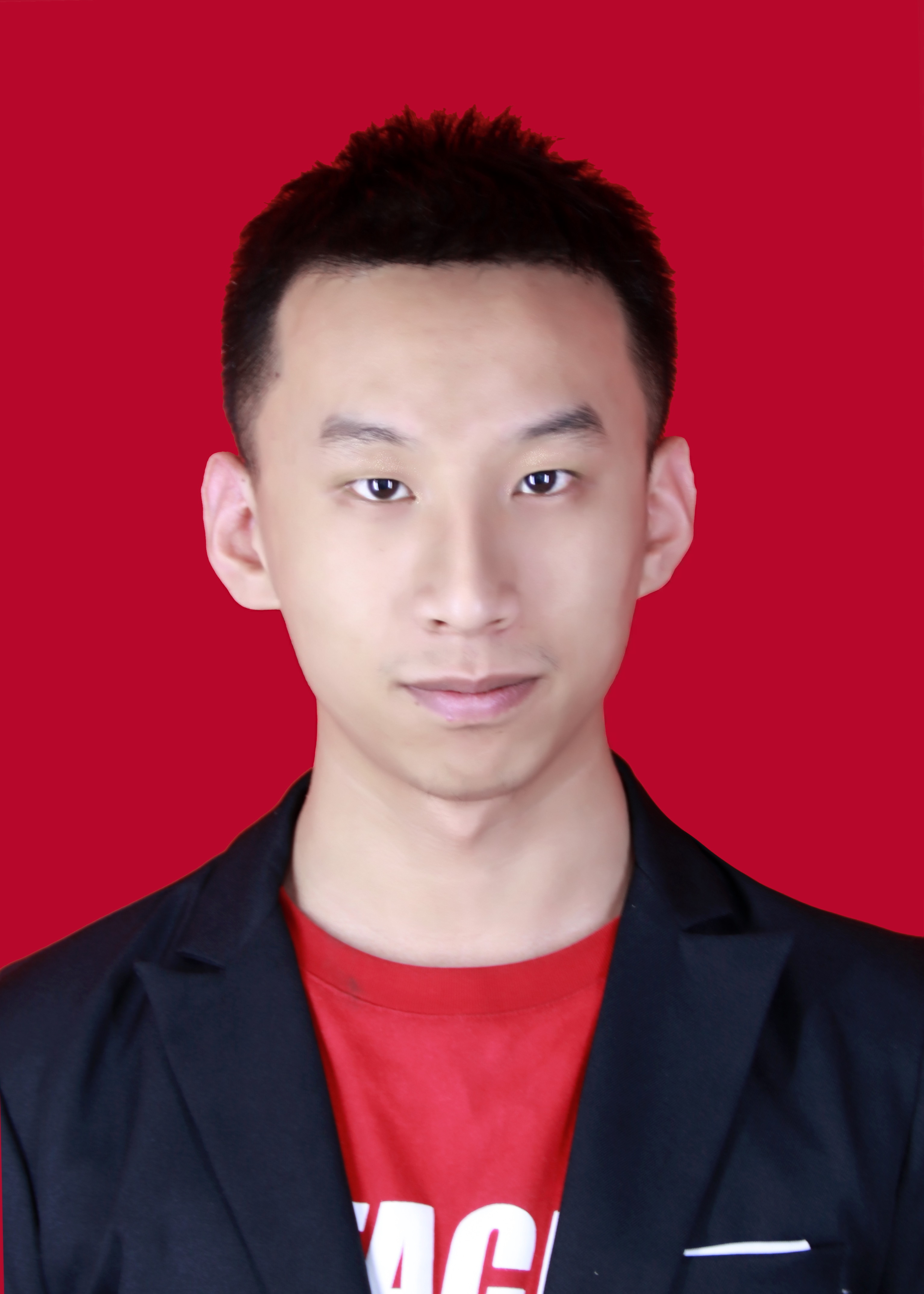}}
	\noindent {\bf Nan Li}\ received the B.Sc. degree in software engineering from the North University of China, Taiyuan, China, in 2020. He is currently pursuing the Ph.D. degree with the College of Software, Northeastern University, Shenyang, China. His research interests include computational intelligence and machine learning.}

\par\noindent 
\parbox[t]{\linewidth}{
	\noindent\parpic{\includegraphics[height=1in,width=0.66in,clip,keepaspectratio]{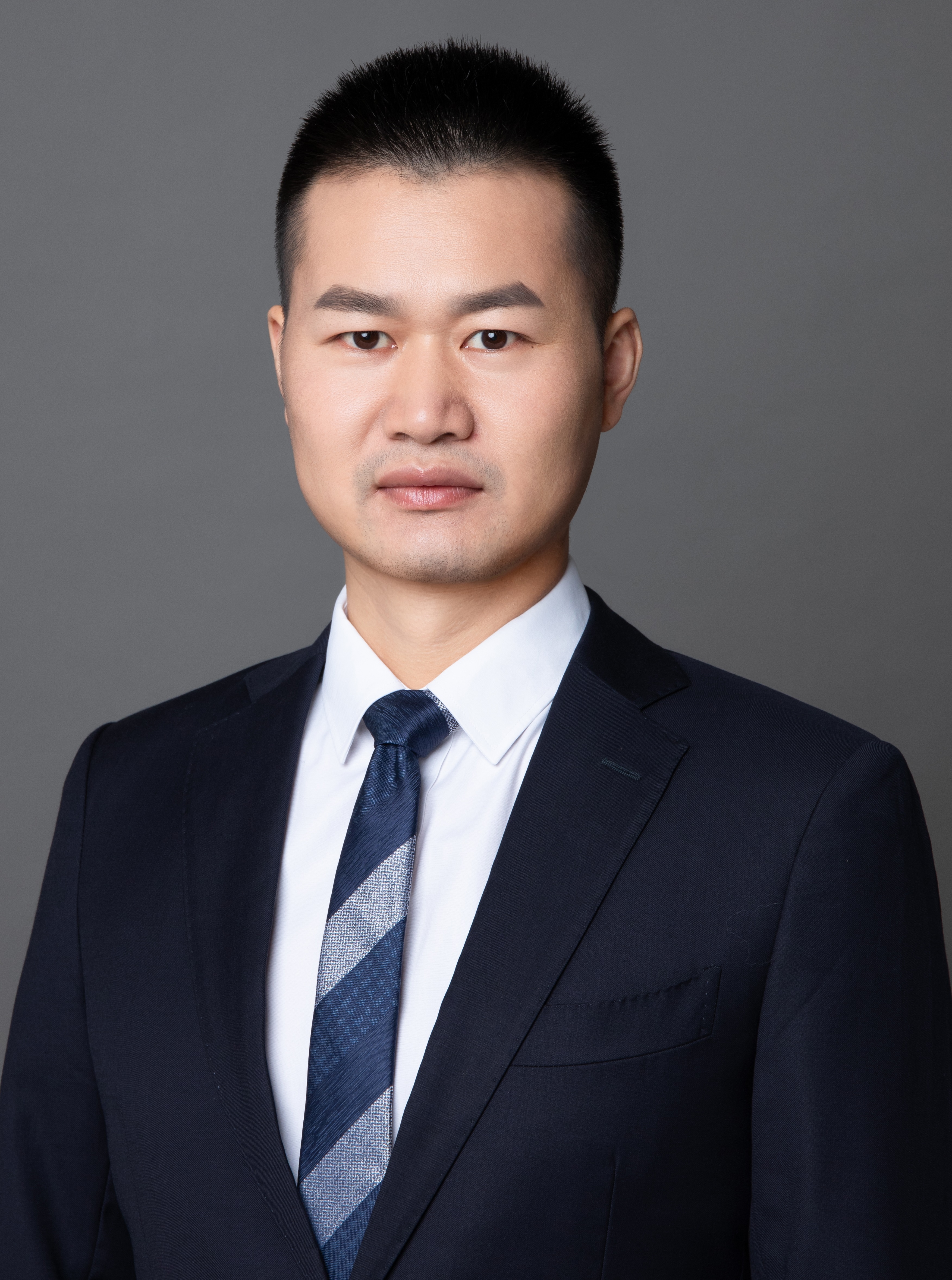}}
	\noindent {\bf Guo Yu}\ (Member, IEEE) received the B.S. degree in information and computing science and the M.Eng. degree in computer technology from Xiangtan University, Xiangtan, China, in 2012 and 2015, respectively. He is currently an associate professor in Nanjing Tech University. His current research interests include evolutionary optimization and deep learning.}

\par\noindent 
\parbox[t]{\linewidth}{
	\noindent\parpic{\includegraphics[height=1in,width=0.66in,clip,keepaspectratio]{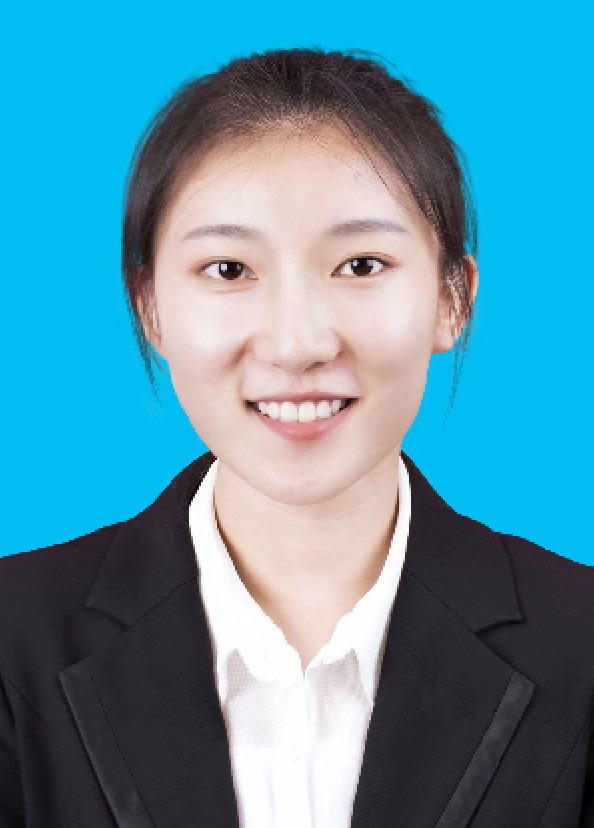}}
	\noindent {\bf Xiaoyu Geng}\ received the B.S. degree in School of Software from Shanxi University, TaiYuan, China and the M.Eng. degree in College of Software from Northeastern University, Shenyang, China, in 2020 and 2023, respectively. Her research interests include computational intelligence and machine learning.}

\par\noindent 
\parbox[t]{\linewidth}{
	\noindent\parpic{\includegraphics[height=1in,width=0.66in,clip,keepaspectratio]{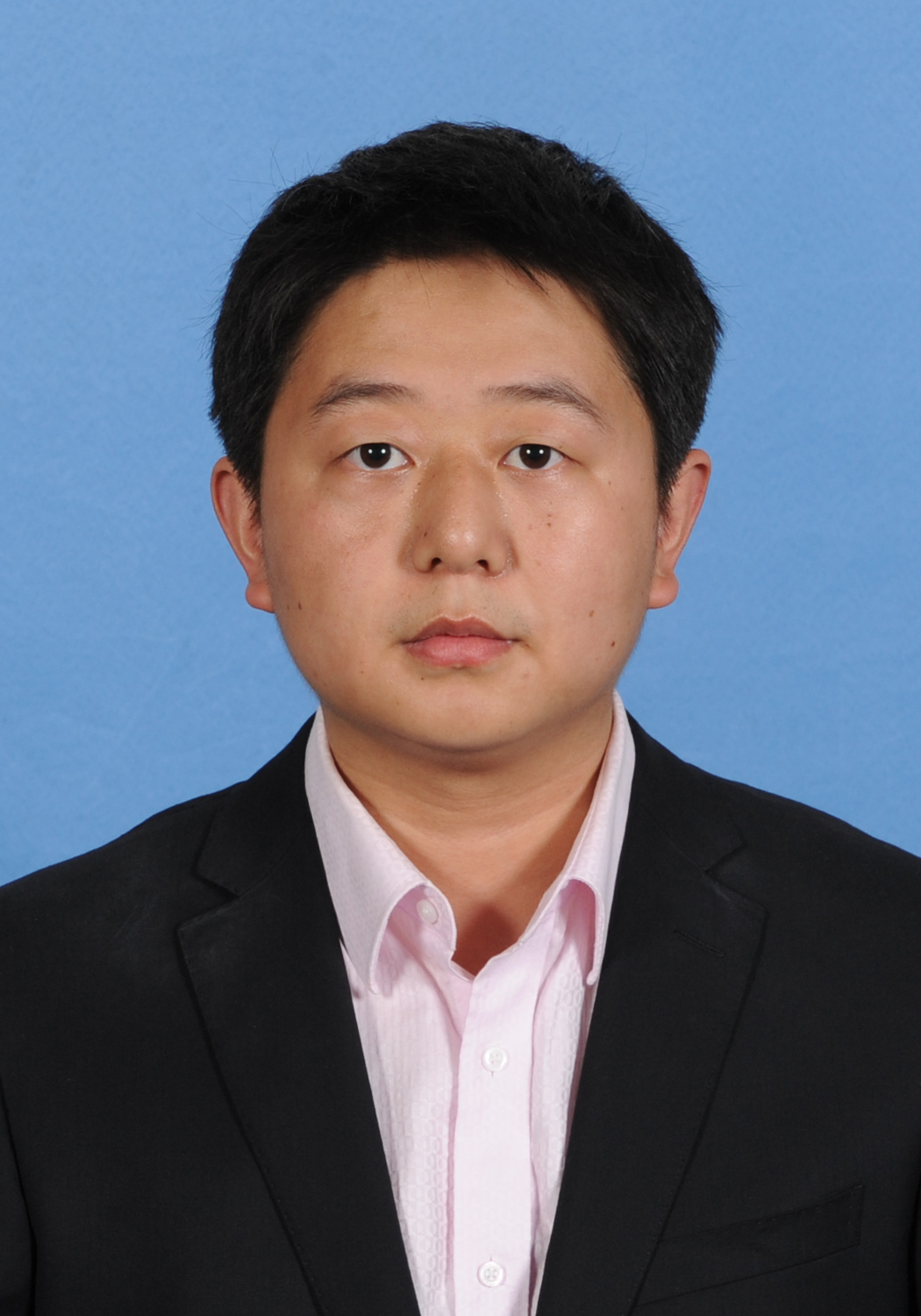}}
	\noindent {\bf Shi Cheng}\ received a bachelor's degree in mechanical and electrical engineering from Xiamen University, Xiamen, China, in 2005, a master's degree in software engineering from Beihang University, Beijing, China, in 2008, a Ph.D. degree in electrical engineering and electronics from the University of Liverpool, Liverpool, U.K. in 2013. His research interests include swarm intelligence, data mining techniques, and their applications.}

\par\noindent 
\parbox[t]{\linewidth}{
	\noindent\parpic{\includegraphics[height=1in,width=0.66in,clip,keepaspectratio]{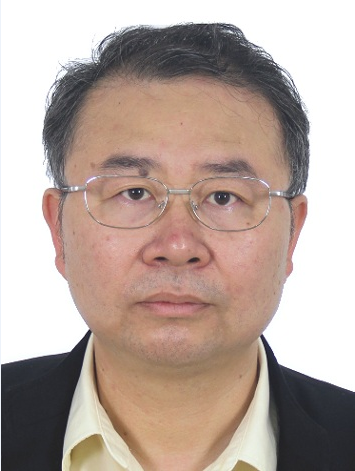}}
	\noindent {\bf Xingwei Wang}\ received the BSc, MSc, and PhD degrees in computer science from the Northeastern University, Shenyang, China, in 1989, 1992, and 1998, respectively. He is currently a professor with the Northeastern University, China. His research interests include cloud computing and future Internet. He has published more than 100 journal articles, books, and refereed conference papers.}

\par\noindent 
\parbox[t]{\linewidth}{
	\noindent\parpic{\includegraphics[height=1in,width=0.66in,clip,keepaspectratio]{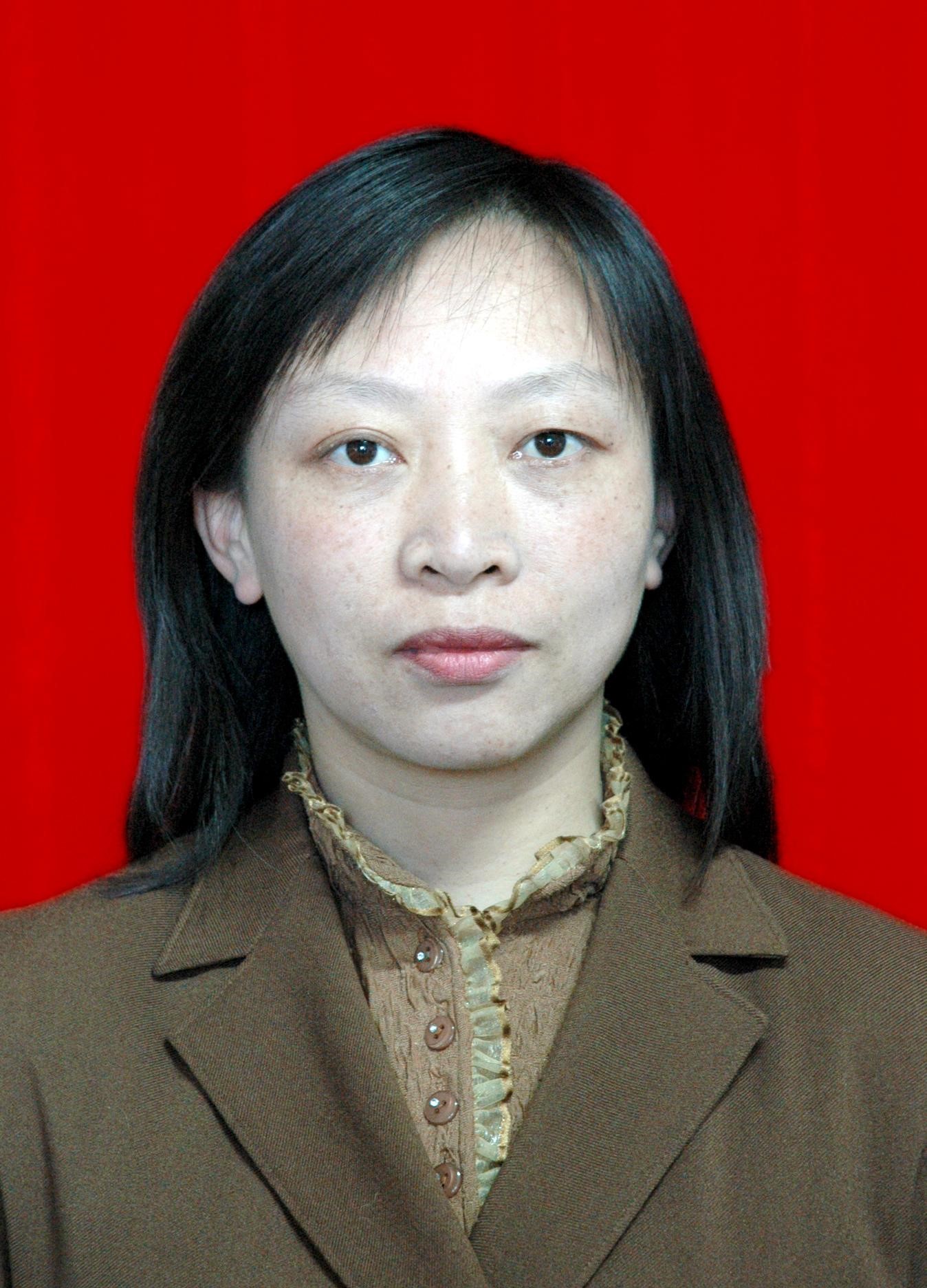}}
	\noindent {\bf Min Huang}\ received the BSc degree in automatic instrument, the MSc degree in systems engineering, and the PhD degree in control theory and control engineering from the Northeastern University, Shenyang, China, in 1990, 1993, and 1999, respectively. She is currently a professor with the Northeastern University, China. Her research interests include modeling and optimization for logistics and supply chain, etc. She has published more than 100 journal articles, books, and refereed conference papers.}

\par\noindent 
\parbox[t]{\linewidth}{
	\noindent\parpic{\includegraphics[height=1in,width=0.66in,clip,keepaspectratio]{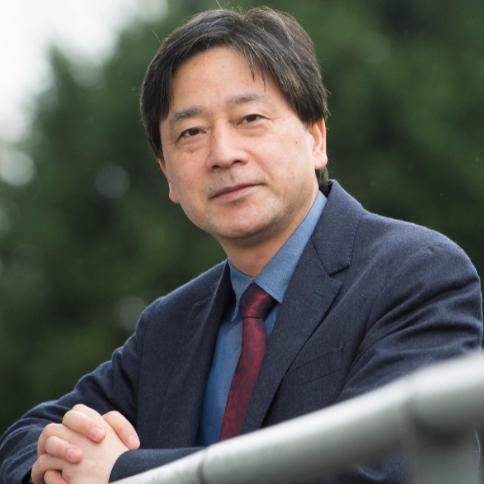}}
	\noindent {\bf Yaochu Jin}\ (Fellow, IEEE) received the B.Sc., M.Sc., and Ph.D. degrees in automatic control from Zhejiang University, Hangzhou, China, in 1988, 1991, and 1996, respectively, and the Dr.-Ing. degree from Ruhr-University Bochum, Bochum, Germany, in 2001. His main research interests include data-driven evolutionary optimization, trustworthy machine learning, multiobjective evolutionary learning, and evolutionary developmental systems. Dr. Jin is the recipient of the 2018 and 2021 IEEE Transactions on Evolutionary Computation Outstanding Paper Award, and the 2015, 2017, and 2020 IEEE Computational Intelligence Magazine Outstanding Paper Award. }

\end{document}